\pdfoutput=1

\documentclass[11pt]{article}

\usepackage[]{EACL2023}

\usepackage{times}
\usepackage{latexsym}
\usepackage{stmaryrd}
\usepackage{amsmath}
\usepackage{multirow}
\usepackage{bbm}
\usepackage{graphicx}
\usepackage{amssymb}
\usepackage{booktabs}
\usepackage{adjustbox}
\usepackage{xspace}
\usepackage{algpseudocode}
\usepackage{algorithm}
\usepackage{graphicx}
\usepackage{caption}
\usepackage{subcaption}
\usepackage{color, colortbl}
\usepackage{tablefootnote}
\usepackage{enumitem}  
\usepackage{lipsum}
\usepackage{xspace}

\usepackage{hyperref}

\def\xHyphenate#1#2\wholeString {\if#1$%
    \else\transform{#1}%
    \takeTheRest#2\ofTheString\fi}
\def\takeTheRest#1\ofTheString\fi
{\fi \xHyphenate#1\wholeString}
\def\transform#1{\url{#1}\hskip 0pt plus 1pt}

\usepackage[T1]{fontenc}

\usepackage[utf8]{inputenc}

\usepackage{microtype}

\newcommand{\datasetname}{NusaWrites}
 
\newcommand{\ex}[1]{\textit{#1}\xspace} 
\newcommand{\equalsign}{\footnotemark[1]\hspace{0.1cm}}

\usepackage{inconsolata}

\setlist{topsep=1pt,itemsep=1pt,partopsep=1pt, parsep=1pt}

\title{NusaWrites: Constructing High-Quality Corpora for\\ Underrepresented and Extremely Low-Resource Languages}

\author{
Samuel Cahyawijaya$^{1,4*}$, Holy Lovenia$^{1,4}$\equalsign, Fajri Koto$^{3,4}\equalsign$, Dea Adhista$^{2\dagger}$, Emmanuel Dave$^{4\dagger}$, \\
\bf{Sarah Oktavianti$^{2\dagger}$, Salsabil Maulana Akbar$^{4\dagger}$, Jhonson Lee$^{4\dagger}$, Nuur Shadieq$^{4\dagger}$,} \\
\bf{Tjeng Wawan Cenggoro$^{8,4\dagger}$, Hanung Wahyuning Linuwih$^{2\dagger}$, Bryan Wilie$^{1,4\dagger}$}, \\
\bf{Galih Pradipta Muridan$^{2\dagger}$, Genta Indra Winata$^{5,4\dagger}$,} \\
\bf{David Moeljadi$^{7\dagger}$, Alham Fikri Aji$^{3,4\dagger}$, Ayu Purwarianti$^{6,4}$, Pascale Fung$^1$} \\
$^1$HKUST \quad $^2$Prosa.ai \quad $^3$MBZUAI \quad $^4$IndoNLP \quad $^5$Bloomberg \\ 
$^6$Institut Teknologi Bandung  $^7$Kanda University of International Studies \quad $^8$Bina Nusantara University \\
$\equalsign$ Equal Contribution \quad $^\dagger$ Equal Contribution
}

\begin{document}
\maketitle
\begin{abstract}

Democratizing access to natural language processing (NLP) technology is crucial, especially for underrepresented and extremely low-resource languages. Previous research has focused on developing labeled and unlabeled corpora for these languages through online scraping and document translation. While these methods have proven effective and cost-efficient, we have identified limitations in the resulting corpora, including a lack of lexical diversity and cultural relevance to local communities. To address this gap, we conduct a case study on Indonesian local languages. We compare the effectiveness of online scraping, human translation, and paragraph writing by native speakers in constructing datasets. Our findings demonstrate that datasets generated through paragraph writing by native speakers exhibit superior quality in terms of lexical diversity and cultural content. In addition, we present the \datasetname{} benchmark, encompassing 12 underrepresented and extremely low-resource languages spoken by millions of individuals in Indonesia. Our empirical experiment results using existing multilingual large language models conclude the need to extend these models to more underrepresented languages. We release the NusaWrites dataset\footnote{Accessible through \texttt{nusacrowd} and HuggingFace \texttt{datasets} packages. Kindly check the README on GitHub for more information.} and code involved in our experiment at \url{https://github.com/IndoNLP/nusa-writes}.

\end{abstract}

\section{Introduction}


Most of the research works in today's NLP technology are culturally Anglocentric with English as the main language~\cite{sogaard2022ban,talat2022reap}. 
While it is critical to democratize NLP to underrepresented languages, previous works~\cite{cahyawijaya2022nusacrowd,kakwani2020indicnlpsuite,koto2020indolem,koto-koto-2020-towards,wilie2020indonlu,tacl_masakhaner,adelani2021masakhaner,cahyawijaya2021indonlg,ebrahimi2022americasnli,park2021klue,kumar2022indicnlg,winata2022nusax,
adilazuarda-etal-2022-indorobusta,
ogundepo2023afriqa,kabra-etal-2023-multi,song2023globalbench} have developed labeled and unlabeled corpora in the languages mainly through document translation~\cite{winata2022nusax} and online scraping~\cite{koto-etal-2021-indobertweet,koto-etal-2022-lipkey}. Although such data collection methods could be effective in high-resource languages, applying the methods in underrepresented languages still needs further investigation.


\begin{figure}[!t]
    \centering
    \includegraphics[width=0.9\linewidth]{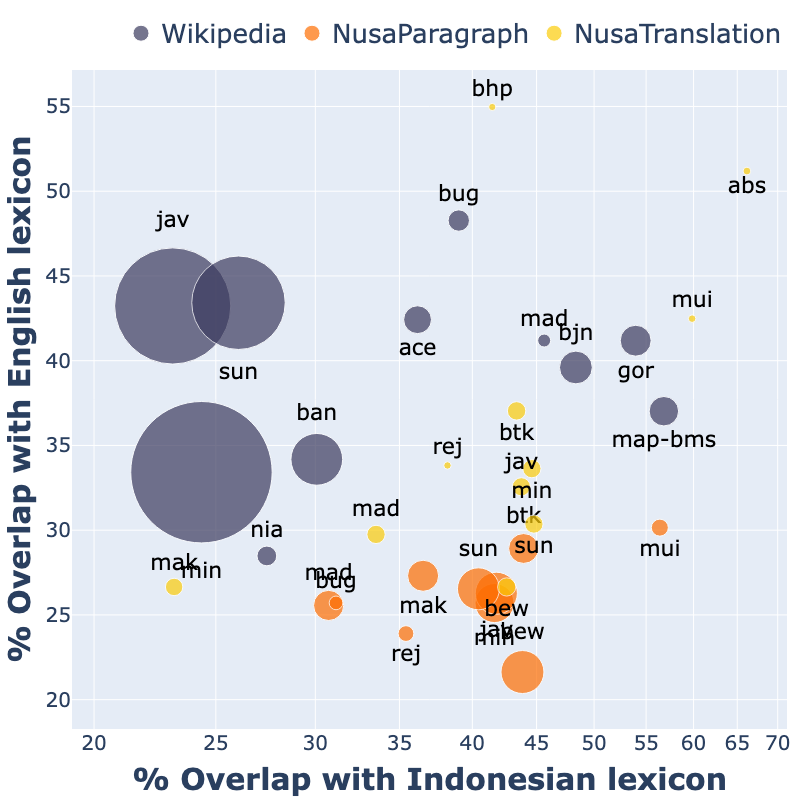} 
     \vspace{-0.7em}
    \caption{Unique lexicon overlaps of various language corpora with Indonesian and English languages. The Indonesian and English lexicons are from Panlex.\protect\footnotemark}
    \label{fig:language-overlap}
     \vspace{-0.5em}
\end{figure}
\footnotetext{\url{https://panlex.org/}}

In this work, we compare three corpus collection methods for 12 underrepresented languages in Indonesia, namely Ambon (abs), Batak (btk), Betawi (bew), Bima (bhp), Buginese (bug), Javanese (jav), Madurese (mad), Makassarese (mak), Minangkabau (min), Palembang / Musi (mui), Rejang (rej), and Sundanese (sun). We chose Indonesian local languages as our case study because of the language diversity in Indonesia, with more than 700 languages spoken but most of them are underrepresented and extremely low-resource~\cite{cohn2014local,aji-etal-2022-one}. \citet{bang2023multitask} categorize Javanese (jav) and Sundanese (sun) as low-resource languages, while the others as extremely low-resource languages. For Ambon (abs), Bima (bhp), Makassarese (mak), Musi (mui), and Rejang (rej), they have no publicly available labeled and unlabeled corpora despite there being millions of speakers. We provide information on 12 low-resource languages under study in Table~\ref{tab:lang-under-study}. We conduct two manual data construction efforts for the 12 languages: topic-focused paragraph writing (NusaParagraph) and human translation by native speakers (NusaTranslation),\footnote{Note that most Indonesians are at least bilingual since they speak Indonesian and their local language~\cite{aji-etal-2022-one,koto-koto-2020-towards}.} and benchmark them with online scraping. For online scraping, we utilize Wikipedia\footnote{\url{https://www.wikipedia.org/}} as the main source as it covers some of the Indonesian local languages under study. Figure~\ref{fig:language-overlap} summarizes the corpora constructed by each approach: Wikipedia, NusaParagraph, and NusaTranslation for online scraping, paragraph writing, and human translation, respectively. NusaParagraph tends to have fewer English and Indonesian lexicons, indicating they are more relevant to the local cultures than the others.



We build a new benchmark for the 12 Indonesian local languages, namely \textbf{\datasetname{}}\footnote{The ``Nusa'' of \datasetname{} is abbreviated from ``Nusantara'', which refers to the Indonesian archipelago.}, using the texts produced in topic-focused paragraph writing and human translation. \datasetname{} covers 5 natural language understanding tasks (e.g., emotion, sentiment classification) and one natural language generation task (i.e., machine translation), and complements NusaX~\cite{winata2022nusax}---a contemporaneous work on 10 Indonesian local languages for sentiment analysis and machine translation.
We also demonstrate the inability of (1) fine-tuned Indonesian and multilingual language models (LMs) and (2) zero-shot prompting via large LMs (LLMs) to adapt to these languages, indicating that these languages are distinct from the existing models.

Our contributions to this work are four-fold:
\begin{itemize}
    \item We compare various corpus collection methods for underrepresented and extremely low-resource languages. We show that paragraph writing is the most promising strategy for building high-quality and culturally-relevant corpora.
    \item We extend the NLP resource coverage of Indonesian local languages with 5 new languages: Ambon (abs), Bima (bhp), Makassarese (mak), Musi (mui), and Rejang (rej). 
    \item We propose \datasetname{},
    a benchmark covering new high-quality human annotated corpora consisting of 12 underrepresented languages in Indonesia with 5 downstream tasks.
    \item We conduct extensive analysis to showcase the similarity between languages under study with Indonesian and the inability of existing LLMs to process these languages.
\end{itemize}

\begin{figure}
    \centering
    \resizebox{0.9\linewidth}{!}{
        \includegraphics{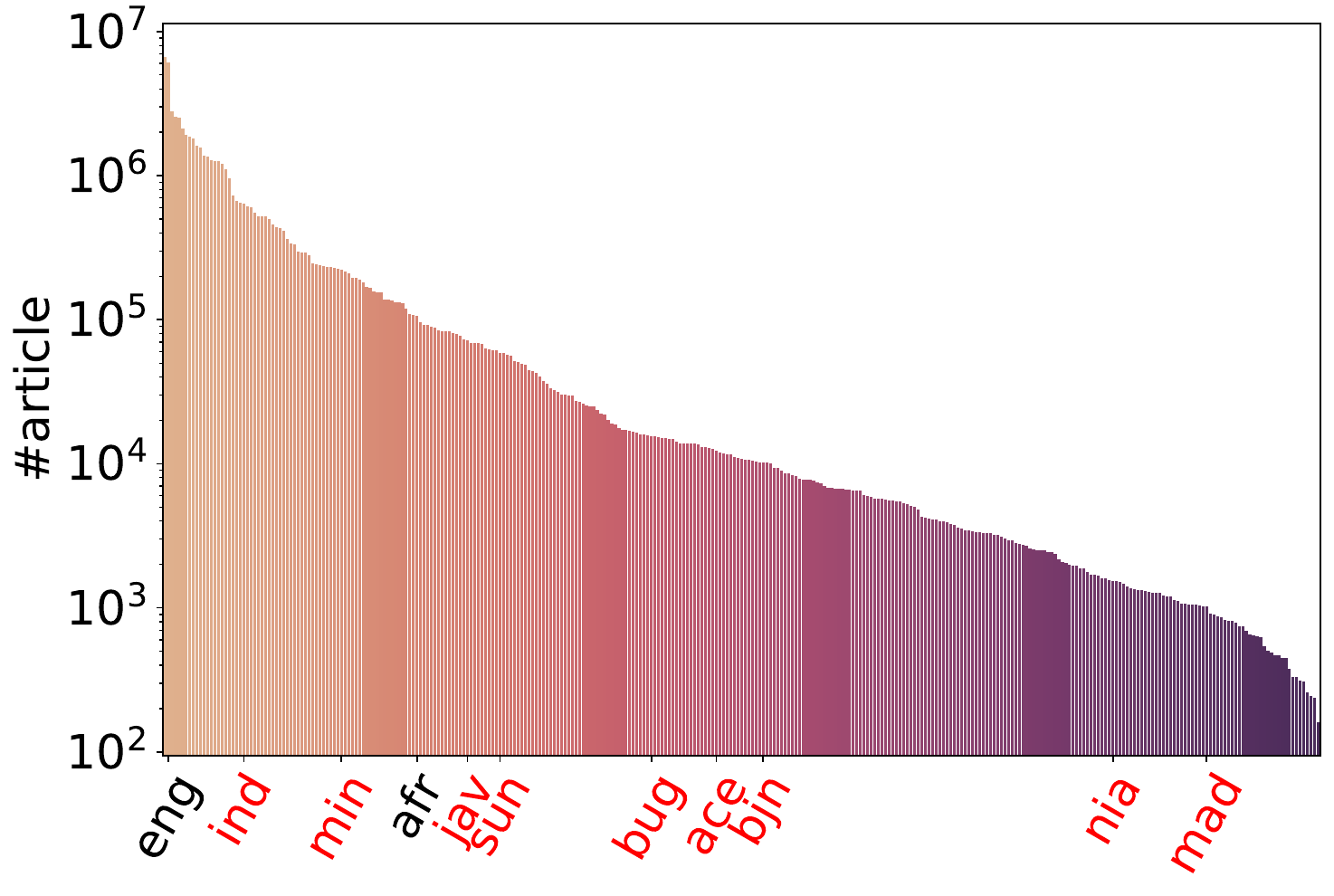}
    }
    \vspace{-0.6em}
    \caption{Distribution of Indonesian languages in Wikipedia, compared against all existing languages.}
    \label{fig:wiki_indo_dist}
     \vspace{-0.7em}
\end{figure}

\section{Indonesian Local Languages in Wikipedia}

Figure~\ref{fig:wiki_indo_dist} describes Indonesian local languages which are covered in Wikipedia, compared against other existing languages. In total, there are only 11 local languages (out of 700+~\cite{aji-etal-2022-one}), with Minangkabau (min), Javanese (jav), and Sundanese (sun) having a quite large amount of documents around $\sim$100,000 articles, while the remaining languages have less than $\sim$10,000 articles. Despite its relatively large scale in Wikipedia, the text quality is not consistently as good as reported in the WikiMatrix dataset~\cite{schwenk2021wikimatrix}. \citet{kreutzer2022qualitymulti} further find that $\sim$30\% of the correct translation data in English-Javanese are either boilerplates or low-quality texts.



To further verify the quality of Indonesian local languages in Wikipedia, we conduct an analysis to measure lexical diversity in two approaches: 1) calculating the cumulative token distribution per language and 2) measuring the length-agnostic lexical diversity metrics, i.e., \textit{moving average type-token ratio} (MATTR)~\cite{covington2010mattr}, \textit{measure of textual lexical diversity} (MTLD)~\cite{mccarthy2005mtld}, and \textit{mean segmental type-token ratio} (MSTTR)~\cite{johnson1944msttr}. We use LexicalRichness~\cite{shen2021accuracybias,shen2022lexicalrichness} v0.5.0\footnote{\url{https://pypi.org/project/lexicalrichness/}} to calculate these metrics. Based on our analysis in Table~\ref{tab:lex-div-wiki}, we show that some Indonesian local languages in Wikipedia have much less lexical diversity despite having quite a number of articles in Wikipedia, especially for Buginese (bug), Acehnese (ace), Gorontalo (gor), and Nias (nia). Through a further inspection of the Wikipedia corpus presented in \S\ref{sec:lexical-div} and Appendix~\ref{app:token-statistics}, Wikipedia articles for these languages tend to comprise many boilerplate texts, especially for Buginese (bug) Wikipedia.



\paragraph{Indonesian Local Languages in Other Sources}

Other than Wikipedia, there are other large multilingual corpora such as CommonCrawl,\footnote{\url{http://commoncrawl.org}} mC4~\cite{xue-etal-2021-mt5}, OSCAR~\cite{ortiz2019oscar}, FLORES-200~\cite{guzman-etal-2019-flores,nllb2022nlbb}, and Bible corpus.\footnote{\url{https://huggingface.co/datasets/bible-nlp/biblenlp-corpus}} Nevertheless, most sources, except the Bible corpus, only support some widely-spoken languages spoken in Indonesia: Indonesian (ind), Javanese (jav), and Sundanese (sun), rendering them ineffective for studying hundreds of local languages spoken in Indonesia. The Bible corpus, on the other hand, consists of 14 Indonesian local languages.\footnote{Details of the Indonesian local languages in the Bible corpus are in Appendix~\ref{app:bible-corpus}. } Interestingly, these languages have an extremely low number of speakers with an average population of 40k people. On the contrary, Wikipedia covers Indonesian local languages with a larger number of speakers, with Nias (nia) being the smallest (nearly 770k speakers). In this work, we particularly focus on Indonesian local languages with larger population size ($\sim$500k or above), and leave the exploration of the smaller-scale languages for future work.

\begin{table}[!t]
    \centering
    \resizebox{\linewidth}{!}{
        \begin{tabular}{c|c|c|c|c|c}
        \toprule
        \textbf{lang} & \textbf{category} & \textbf{MATTR} & \textbf{MTLD} & \textbf{MSTTR} & \textbf{Avg.} \\
        \midrule
        \textbf{gor} & \textbf{X-LRL} & 69.40 & 37.23 & 71.36 & \textbf{47.04} \\
        \textbf{ace} & \textbf{X-LRL} & 77.87 & 30.65 & 75.91 & \textbf{51.36} \\
        \textbf{bug} & \textbf{X-LRL} & 79.81 & 28.61 & 80.12 & \textbf{53.41} \\
        \textbf{nia} & \textbf{X-LRL} & 84.75 & 68.85 & 86.33 & \textbf{57.25} \\
        \textbf{ban} & \textbf{X-LRL} & 85.15 & 53.83 & 86.57 & \textbf{57.42} \\
        \textbf{map-bms} & \textbf{X-LRL} & 86.62 & 70.76 & 87.89 & \textbf{58.41} \\
        \textbf{bjn} & \textbf{X-LRL} & 87.27 & 83.57 & 88.20 & \textbf{58.77} \\
        \textbf{jav} & \textbf{LRL} & 89.18 & 58.94 & 88.19 & \textbf{59.32} \\
        \textbf{ind} & \textbf{MRL} & 89.88 & 83.82 & 90.11 & \textbf{60.27} \\
        \textbf{mad} & \textbf{X-LRL} & 89.88 & 67.21 & 90.53 & \textbf{60.36} \\
        \textbf{sun} & \textbf{LRL} & 94.47 & 70.12 & 88.92 & \textbf{61.37} \\
        \textbf{min} & \textbf{X-LRL} & 94.23 & 80.86 & 92.12 & \textbf{62.39} \\
        \bottomrule
        \end{tabular}
    }
    \caption{Lexical diversity of various Indonesian local languages corpora in Wikipedia. \textbf{X-LRL} = Extremely low-resource language, \textbf{LRL} = low-resource language, and \textbf{MRL} = medium-resource language.}
    \label{tab:lex-div-wiki}
\end{table}

\section{Corpus Construction for Indonesian Local Languages}

We conduct corpus construction through human annotation by expert workers in two ways: (1) sentence translation and (2) paragraph writing. Sentence translation is a widely used parallel data collection method~\cite{conneau2018xnli,Hu2020xtreme,winata2022nusax}, while paragraph writing~\cite{koto-etal-2022-cloze} is explored to capture a more culturally relevant aspect which is often left out in translation~\cite{kirkpatrick2009lost}. 
The details of our expert annotator recruitment are shown in Appendix~\ref{app:pre-annotation}. 
In the following section, we describe how the data construction is done for both methods.

\subsection{Sentence Translation}

\paragraph{Data Selection}
We sample data from two sources, i.e., IndoLEM sentiment~\cite{koto2017inset, koto2020indolem}, an Indonesian sentiment analysis dataset collected from Twitter and hotel review, and EmoT~\cite{saputri2018emotion, wilie2020indonlu}, an Indonesian emotion classification dataset collected from Twitter. We take the whole samples 
from both IndoLEM sentiment (5048 samples) and EmoT (4401 samples) as our source language data, resulting in a total data of 9,449 sentences for translation.

\paragraph{Translation Procedure} We translate the source language data into 11 languages: Ambon (abs), Batak (btk), Betawi (bew), Bima (bhp), Javanese (jav), Madurese (mad), Makassarese (mak), Minangkabau	(min), Musi (mui), Rejang (rej), and Sundanese (sun). 
Our expert annotators are instructed to translate while maintaining: (1) the sentence's sentiment/emotion polarity; (2) the named entities; and (3) the completeness of the text. The translation procedure is detailed in Appendix~\ref{app:translation-proc}. 

\subsection{Paragraph Writing}

We conduct paragraph writing by instructing the annotators to write a 100-word paragraph given a certain topic. The topic for paragraph writing is manually designed to cover a wide coverage of domains. We conduct paragraph writing in 10 languages, i.e., Batak (btk), Betawi (bew), Buginese (bug), Javanese (jav), Madurese (mad), Makassarese (mak), Minangkabau (min), Musi (mui), Rejang (rej), and Sundanese (sun). Note that, unlike sentence translation, there is no Ambon (abs) and Bima (bhp), but instead there is an additional language, Buginese (bug). This happens because of the difficulty of obtaining a large pool of annotators for the Ambon (abs) and Bima (bhp) languages.

\paragraph{Topic Selection} 
We provide a list of topics before instructing the annotators to write paragraphs. The selection of various topics is expected to enrich vocabulary in the corpus as different topics will obviously bring up the use of different terms. The topics provided vary widely, ranging from food and beverages, entertainment/leisure, sports, science, history, politics, and religion. In addition, we also have other topics for describing emotional states such as sadness, happiness, anger, etc. We also provide more specific subtopics for each of the major topics provided. In total, we have 20 main topics with 20 subtopics of each main topic. The list of all topics is given in Appendix~\ref{app:topic-selection}. 

\paragraph{Paragraph Writing Procedure}

The paragraph writing is done with the following criteria: (a) the paragraph consists of a minimum of 100 words, (b) using only the targeted local language except for named entities, (c) the content of the paragraph should be about the provided topics and subtopics, and (d) for each paragraph, the annotator should fill the rhetoric type of the paragraph, which is either narration, description, argumentation, persuasion, or exposition. More details about the paragraph writing procedure are in Appendix~\ref{app:paragraph-writing}.

\subsection{Quality Control}

Quality control is conducted to ensure the data are correct through manual and automatic validation. If the data does not meet the desired criteria, it is to be revised. Specifically, through a series of manual and automatic validations, we ensure that all sentences that need to be translated are translated to the target language, with minimal overlap with the source language sentence. For paragraph writing, we ensure that there is no plagiarism from external sources by conducting validation through search engines and we also ensure that there is a minimum 30\% distinction between two paragraphs (measured by using edit distance). The detail of our quality control process is described in Appendix~\ref{app:data-validation}. The quality control is conducted in several iterations, by asking annotators to rewrite unqualified instances until all quality control passes.

\begin{figure*}[t]
     \centering
     \begin{subfigure}[b]{0.29\textwidth}
         \centering
         \includegraphics[width=\textwidth]{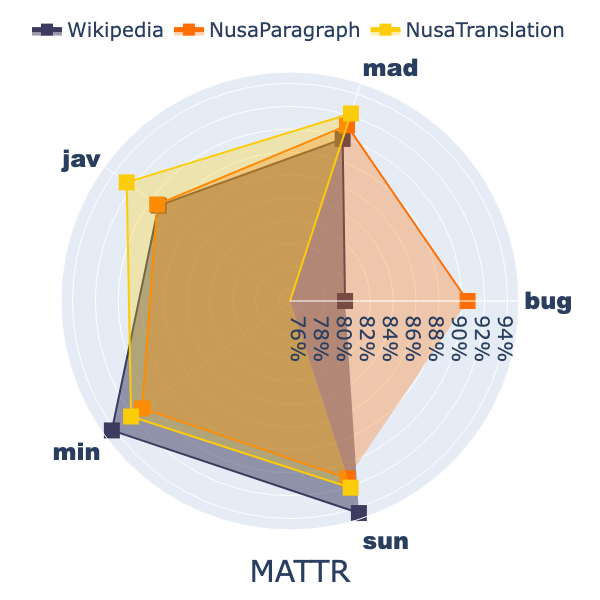}
     \end{subfigure}
     \begin{subfigure}[b]{0.29\textwidth}
         \centering
         \includegraphics[width=\textwidth]{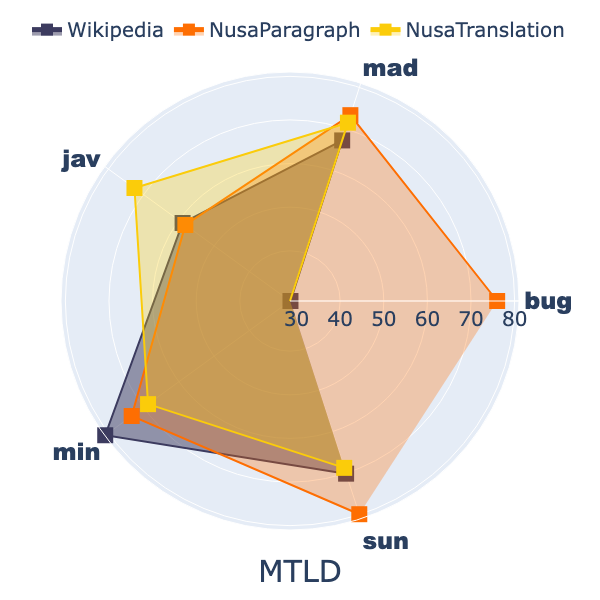}
     \end{subfigure}
     \begin{subfigure}[b]{0.29\textwidth}
         \centering
         \includegraphics[width=\textwidth]{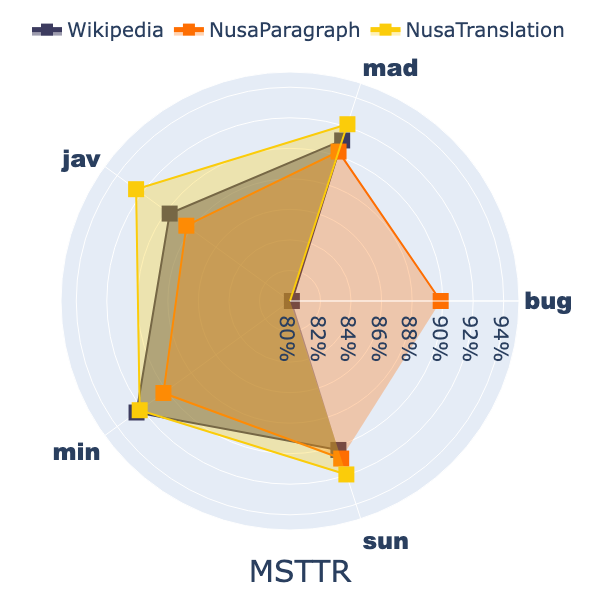}
     \end{subfigure}
    \vspace{-0.6em}
    \caption{The \textbf{(left)} MATTR, \textbf{(center)} MTLD, and \textbf{(right)} MSTTR scores of different corpus collection methods. Paragraph writing and translation achieve higher diversity on the extremely low-resource languages, i.e., Madurese (mad) and Buginese (bug), compared to scraping from Wikipedia.}
    \label{fig:lexical-diversity}
     \vspace{-0.5em}
\end{figure*}

\subsection{Resulting Corpora}
\label{sec:resulting-corpora}

Through sentence translation, we achieve a total of 72,444 sentences, with 1,579 for Bima; 1,574 each for Musi, Rejang, and Ambon; and 9,449 each for Madurese, Minangkabau, Batak, Betawi, Javanese, Sundanese, and Makassarese. As for paragraph writing, we achieve a total of 57,409 paragraphs, specifically: 5,211 for Maduranese, 8,608 for Minangkabau, 10,188 for Javanese, 9,594 for Sundanese, 9,755 for Betawi, 4,908 for Batak, 5,471 for Makassar, 1,200 for Rejang, 1,474 for Musi, and 1,000 for Buginese. We develop two corpora grouped according to how the data is constructed: \textbf{NusaTranslation} and \textbf{NusaParagraph}.
We use the NusaTranslation and NusaParagraph corpora to build \datasetname, an underrepresented language benchmark for 12 Indonesian local languages.


\section{\datasetname{} over Wikipedia}
\label{sec:writing-over-wiki}

In this section, we compare the quality of various corpus collection methods, i.e., online scraping from Wikipedia, sentence translation (NusaTranslation), and paragraph writing (NusaParagraph), through 5 Indonesian local languages: Buginese (bug), Javanese (jav), Madurese (mad), Minangkabau (min), and Sundanese (sun).

The statistics of each corpus collection method as shown in Appendix~\ref{app:token-statistics}. In general, Wikipedia has a larger token count and unique token coverage for Javanese (jav), Sundanese (sun), and Minangkabau (min). While for Madurese (mad), the corpus size in Wikipedia is very small with only 110k tokens, in this case, the sentence translation and paragraph writing methods provide a huge advantage over collecting through Wikipedia. Interestingly, while the \#tokens of the Buginese (bug) in Wikipedia are rather large, the \#unique tokens are very small even compared to the smaller data from Madurese (mad). Additionally, the \#tokens/document is pretty small, indicating a short document per Wikipedia article. These facts show that the data for Buginese (bug) in Wikipedia comprises many short boilerplate texts, which are not useful for learning the language. 

As the data size for each corpus collection method is different, to further compare the corpus quality generated by each corpus collection method, we compare three criteria that are less prone to the size of the data, i.e., length-agnostic lexical diversity metrics; the empirical language modeling quality from LMs trained on the generated corpus on hold-out text data, NusaX~\cite{winata2022nusax}, a human-translated Indonesian social media posts and online reviews; and the ratio of borrowed words of the generated corpus.

\subsection{Lexical Diversity}
\label{sec:lexical-div}

To measure the lexical diversity, we measure the length-agnostic lexical diversity metrics, i.e.,~MATTR~\cite{covington2010mattr}, MTLD~\cite{mccarthy2005mtld}, and MSTTR~\cite{johnson1944msttr}, for each corpus collection method in Figure~\ref{fig:lexical-diversity}.\footnote{We zero out the Buginese (bug) statistics for the sentence translation as we do not collect Buginese (bug) data in the sentence translation.} For MTLD we use a threshold of 0.72, while for  MATTR and MSTTR, we use a window size of 20.\footnote{According to \cite{lembersky2013improving}, lexical diversity through TTR might be inherently different for translated texts which might affect the result from NusaTranslation.} For low-resource languages, i.e., Javanese (jav) and Sundanese (sun), all three methods produce an almost equally diverse corpus, with a slightly higher diversity for sentence translation. For extremely low-resource languages, compared to other methods, Wikipedia achieves slightly higher diversity scores in Minangkabau (min), and NusaTranslation achieves slightly higher scores in Madurese (mad). Nonetheless, by utilizing permutation test ($n=1,000$)~\cite{koplenig2019permutationtest}, we conclude that the difference between corpora in all metrics and languages are statistically significant ($p<0.05$), except for Madurese (mad) between NusaTranslation and NusaParagraph on the MATTR and MTLD metrics, and for Sundanese (sun) between NusaTranslation and NusaParagraph on the MATTR and MSTTR metrics. Interestingly, for Buginese (bug), Wikipedia achieves very low score diversity scores, while NusaParagraph achieves high diversity scores, which shows that there are a large number of sentences in the  Wikipedia data for Buginese (bug) that have a repeating pattern like boilerplate.
In addition to the diversity metrics, we also measure the lexical overlapping with Indonesian and English lexicons obtained from Panlex~\cite{kamholz2014panlex}. As shown in Figure~\ref{fig:language-overlap}, Wikipedia has a higher overlap with the English lexicon, indicating that it covers many shared foreign terms (e.g., scientific terms) and foreign entities (e.g., the name of cities, tourist attractions, etc.), which are not common in the actual day-to-day use of Indonesian local languages where the languages are commonly used for daily conversation, instead of a formal occasion, such as in the academic setting~\cite{cohn2014local, soeparno2015kerancuan,nurjanah2018pengembangan,apriani2018realisasi,sutrisno2019beyond}.

\subsection{Language Modeling Quality}

\begin{figure}[!t]
     \centering
     \begin{subfigure}[b]{0.545\linewidth}
         \centering
         \includegraphics[width=\linewidth]{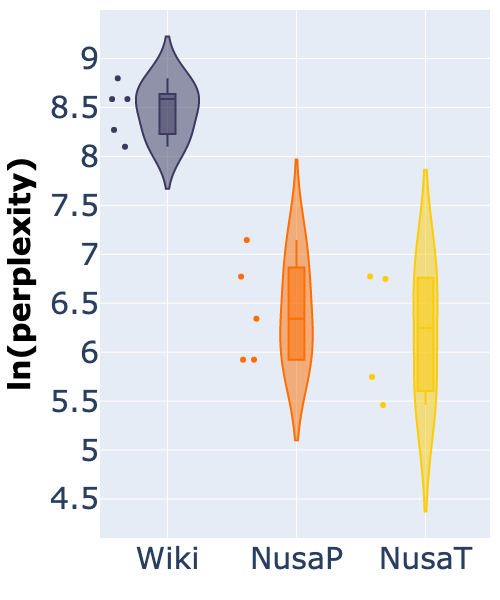}
         \vspace{-2.0em}
         \caption{\textbf{Balanced} setting}
         \label{fig:balanced-ppl}
     \end{subfigure}
     \begin{subfigure}[b]{0.435\linewidth}
         \centering
         \includegraphics[width=\linewidth]{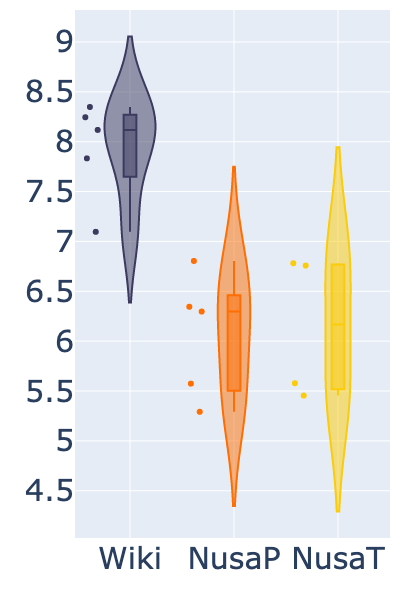}
         \vspace{-2.0em}
         \caption{\textbf{Full} setting}
         \label{fig:full-ppl}
     \end{subfigure}
    \vspace{-1.5em}
    \caption{LMs perplexity evaluation of different corpus collection methods. Lower is better. Wiki: Wikipedia, NusaP: NusaParagraph, NusaT: NusaTranslation.}
\end{figure}

To evaluate the quality of the generated corpus, we evaluate the LM trained on each of the corpus generated by each method. Specifically, we build a small two-layer decoder-only transformer with 128 hidden dimensions and total parameters of $\sim$5.5M, which is a comparable size to a BERT-Tiny model~\cite{devlin2019bert} using two different settings: 1) using the same amount of tokens for each corpus by sampling the larger-sized corpora (\textbf{balanced}) and 2) using the original corpus size for each collection method (\textbf{full}).\footnote{All models are trained using the IndoGPT tokenizer (\url{https://huggingface.co/indobenchmark/indogpt}).} The first one shows the expected quality of the sentences in the corpora, while the second one shows the expected empirical performance when utilizing the corpus.

The LM perplexity of the three corpus collection methods is shown in Figure~\ref{fig:balanced-ppl}. In general, the performance of LMs from NusaTranslation and NusaParagraph is much lower than the one from Wikipedia, showing that the corpora are more aligned with the colloquial writing of Indonesian local languages which is the common use case of using these languages~\cite{cohn2014local,farisiyah2018local,soeparno2015kerancuan,nurjanah2018pengembangan,apriani2018realisasi,sutrisno2019beyond,aji-etal-2022-one}. For the \textbf{balanced} setting, we observe that LMs from NusaTranslation produce slightly better results than the LMs from NusaParagraph. This is expected as the source domain of NusaTranslation is more similar to NusaX~\cite{winata2022nusax}, which also covers social media content and online reviews. Nevertheless, as shown in the results from the \textbf{full} setting (see Figure~\ref{fig:full-ppl}), this problem can be alleviated by increasing the coverage of the corpus.

\subsection{Loan Words Ratio}

\begin{figure}
    \centering
    \resizebox{\linewidth}{!}{
        \includegraphics{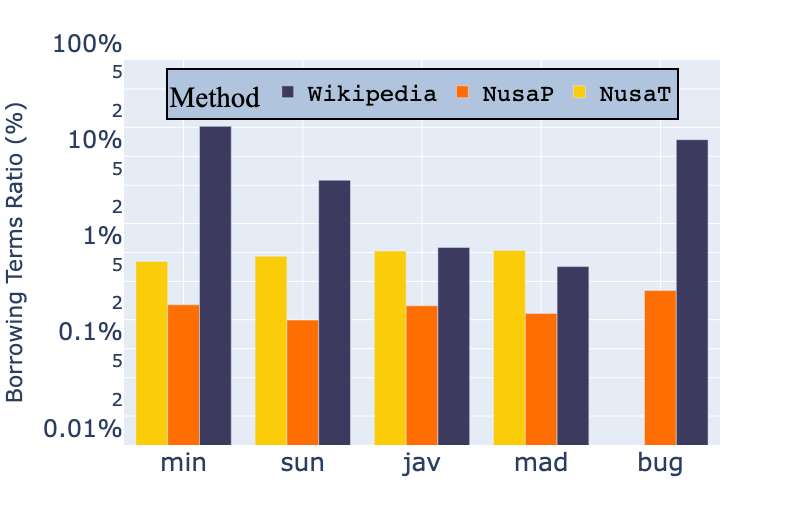}
    }
    \vspace{-1.8em}
    \caption{Ratio of loan words per language of different corpus collection methods. Wiki: Wikipedia, NusaP: NusaParagraph, NusaT: NusaTranslation. The ratio is presented in log$_{10}$ basis.}
    \label{fig:borrowing-term}
\end{figure}


\begin{table*}[t]
\begin{subtable}[!t]{0.635\textwidth}
\centering
\resizebox{0.95\linewidth}{!}{%
\begin{tabular}{lccccc}
\toprule
 & \multicolumn{3}{c}{\textbf{NusaParagraph}} & \multicolumn{2}{c}{\textbf{NusaTranslation}} \\ \cmidrule(lr){2-4} \cmidrule(lr){5-6}
\multirow{-2}{*}{\textbf{Models}} & \textbf{Emot} & \textbf{Rhetorical Mode} & \textbf{Topic} & \textbf{Emot} & \textbf{Senti} \\ \midrule
\multicolumn{6}{c}{\cellcolor[HTML]{D9D9D9}\textit{\textbf{Classical}}} \\
Logistic Regression & \underline{\textbf{78.23}} & 45.21 & \underline{\textbf{87.67}} & \textbf{56.18} & 74.89 \\
Naive Bayes & 75.51 & 37.73 & 85.06 & 52.70 & 74.89 \\
SVM & 76.36 & \textbf{45.44} & 85.86 & 55.08 & \textbf{76.04} \\ \midrule
\multicolumn{6}{c}{\cellcolor[HTML]{D9D9D9}\textit{\textbf{Fine-tuning}}} \\
IndoLEM IndoBERT$_\text{BASE}$ & 66.94 & \underline{\textbf{51.93}} & 84.87 & 52.59 & 69.08 \\
IndoNLU IndoBERT$_\text{BASE}$ & 67.12 & 47.92 & \textbf{85.87} & 54.50 & 75.24 \\
IndoNLU IndoBERT$_\text{LARGE}$ & 62.65 & 31.75 & 85.41 & \underline{\textbf{57.80}} & 77.40 \\
mBERT & 63.15 & 50.01 & 73.82 & 44.13 & 68.72 \\
XLM-R$_\text{BASE}$ & 59.15 & 49.17 & 71.68 & 47.02 & 68.62 \\
XLM-R$_\text{LARGE}$ & \textbf{67.42} & 51.57 & 83.05 & 54.84 & \underline{\textbf{79.06}} \\ \midrule
\multicolumn{6}{c}{\cellcolor[HTML]{D9D9D9}\textit{\textbf{Zero-shot}}} \\
BLOOMZ-560M & 6.57 & 11.60 & 4.98 & 14.27 & 46.22 \\
BLOOMZ-1.1B & 11.72 & \textbf{12.76} & 5.28 & 13.64 & 60.64 \\
BLOOMZ-1.7B & 8.56 & 9.92 &	12.73 &	11.77 &	\textbf{65.10} \\
BLOOMZ-3B & 8.54 & 12.35 & 13.55 &	16.03 &	62.23 \\
BLOOMZ-7.1B & 13.87 & 10.04 & 11.16 & 11.05 & 57.84 \\
mT0$_\text{SMALL}$ & 9.35 & 8.61 & 32.04 & 15.33 & 31.92 \\
mT0$_\text{BASE}$ & 13.76 & 7.70 & 35.35 & 23.56 & 27.70 \\
mT0$_\text{LARGE}$ & 12.18 & 7.50 & 31.00 & 21.91 & 35.25 \\
mT0$_\text{XL}$ & \textbf{21.97} & 8.70 & 31.76 & \textbf{30.36} & 40.11 \\
mT0$_\text{XXL}$ & 19.08 & 8.48 & \textbf{40.22} & 23.49 & 35.44 \\
\bottomrule
\end{tabular}%
}
\caption{NLU evaluation results of \datasetname{}.}
\label{tab:nlu_results}
\end{subtable}
\begin{subtable}[t]{0.35\textwidth}
\centering
\resizebox{0.955\linewidth}{!}{%
\begin{tabular}{lcc}
\toprule
&& \\
\multirow{-2}{*}{\textbf{Models}}
 & \multirow{-2}{*}{\textbf{SacreBLEU}} & \multirow{-2}{*}{\textbf{ChrF++}} \\ \midrule
\multicolumn{3}{c}{\cellcolor[HTML]{D9D9D9}\textit{\textbf{Classical}}}\\
Copy & 23.49 & 41.90 \\
Word Substitution & \textbf{23.80} & 42.68 \\
PBSMT & 25.00 & \underline{\textbf{56.60}} \\ \midrule
\multicolumn{3}{c}{\cellcolor[HTML]{D9D9D9}\textit{\textbf{Fine-tuning}}} \\
IndoBART & \underline{\textbf{30.88}} & \textbf{51.09} \\
IndoGPT & 27.36 & 49.25 \\
mBART-50 & 23.40 & 40.32 \\
mT5 & 26.16 & 46.84 \\
\midrule
\multicolumn{3}{c}{\cellcolor[HTML]{D9D9D9}\textit{\textbf{Zero-shot}}} \\
BLOOMZ-560M & 3.14 & 18.90 \\
BLOOMZ-1.1B & 2.12 & 16.36 \\
BLOOMZ-1.7B & 4.68 & 21.70 \\
BLOOMZ-3B & 5.70 & \textbf{24.34}\\
BLOOMZ-7.1B & 3.65 & 12.42 \\
mT0$_\text{SMALL}$ & 2.35 & 11.82 \\
mT0$_\text{BASE}$ & 3.14 & 13.28 \\
mT0$_\text{LARGE}$ & 2.39 & 11.29 \\
mT0$_\text{XL}$ & 4.22 & 16.13 \\
mT0$_\text{XXL}$ & \textbf{6.33 } & 16.15 \\
\bottomrule
\end{tabular}%
}

\caption{NLG evaluation results of \datasetname{}.}
\label{tab:nlg_results}
\end{subtable}
\caption{Overall performance on all tasks in the \datasetname{} benchmark. We report the macro-F1 (\%) for NLU, and SacreBLEU and ChrF++ for NLG, averaged over all of the languages within the tasks. The best performances in each section are \textbf{bolded}, while the best overall performance in each column is \underline{underlined}.}
\end{table*}

To assess the cultural relevance of the generated corpora, we evaluate the ratio of loan words present within each corpus. The loan words are manually curated from the top 200 words that overlap with the English lexicon and an additional list of English loan words\footnote{The English loan words for local languages are commonly shared with the English loan words in Indonesian. The list of English loan words is collected from \url{https://id.wiktionary.org/wiki/Wikikamus:ProyekWiki_bahasa_Indonesia/Daftar_kata/Serapan/Inggris}.} in each corpus.\footnote{We use the English lexicon instead of Indonesian because decoupling the word borrowing from Indonesian is impractical due to the relatively high terms overlapping coming from the shared geopolitical landscape and cultural values.} The complete list of loan words is in Appendix~\ref{app:borrowed-words}. The ratio is calculated by dividing the number of loan words by the total number of tokens in each corpus, and the results are presented in Figure~\ref{fig:borrowing-term}. The findings indicate that NusaParagraph and NusaTranslation exhibit a minimal ratio of loan words, with approximately $\sim$0.1\% and $\sim$1\% respectively. However, some languages in Wikipedia, such as Minangkabau (min), Sundanese (sun), and Buginese (bug), demonstrate significantly higher ratios of loan words, ranging from approximately 5\% to 15\%. Additionally, in Appendix~\ref{app:common-word-ratio}, we demonstrate that NusaParagraph and NusaTranslation possess a notably higher ratio of common local words, including terms like \texttt{indomie} and \texttt{angkot}, in comparison to Wikipedia. These results emphasize the superiority of manually curated methods, particularly paragraph writing, in generating culturally relevant corpora.

\section{\datasetname{} Benchmark}
\label{sec:benchmark}


From our resulting corpora in \S\ref{sec:resulting-corpora}, we build the \datasetname{} benchmark, which consists of 12 Indonesian local languages: Ambon (abs), Batak (btk), Betawi (bew), Bima (bhp), Buginese (bug), Javanese (jav), Madurese (mad), Makassarese (mak), Minangkabau (min), Palembang / Musi (mui), Rejang (rej), and Sundanese (sun). More details of each language are in Appendix~\ref{app:lang_study}. 4 languages under study, i.e.,  Ambon (abs), Bima (bhp), Musi (mui), and Rejang (rej), have a population of <1M speakers, while others have a population of >2M speakers, but are underrepresented in NLP research~\cite{van2022writing,aji-etal-2022-one}.
The languages belong to the Austronesian language family under the Malayo-Polynesian subgroup. 
While some of the languages are written in multiple scripts, we use the Latin script in \datasetname{}, which has become predominant for all covered languages.


\subsection{NusaTranslation}

We develop three parallel downstream tasks---sentiment analysis, emotion recognition, and machine translation---covering 11 local languages spoken in Indonesia. We generate a new split for each downstream task and keep a reasonable amount of test samples for languages with smaller sample sizes. The labels of the downstream tasks follow the original label from the original dataset. The statistics of each downstream task are shown in Table~\ref{tab:nusatranslation_stats}. A detailed description of each downstream task is provided in Appendix~\ref{app:nusa_translation}. 

\subsection{NusaParagraph}

We develop three downstream tasks from NusaParagraph---topic modeling, emotion recognition, and rhetoric mode classification---based on the datasets covering 10 local languages spoken in Indonesia.
For the topic modeling task, we cover 8 topics: food \& beverages, sports, leisure, religion, culture \& heritage, a slice of life, technology, and business. For the emotion recognition task, we cover the 6 basic emotions~\cite{ekman1992basicemotion}: fear, disgusted, sad, happy, angry, and surprise, and an additional emotion label: shame~\cite{poulson2000shame}. For the rhetoric mode classification, we cover 5 rhetoric modes: narrative, persuasive, argumentative, descriptive, and expository. The statistics of the corpus and the detailed description of each task are shown in Table~\ref{tab:nusaparagraph_stats} and Appendix~\ref{app:nusa_writing}.




\subsection{Baselines}

\paragraph{Classical Machine Learning}

In extremely low-resource settings, the classical approaches can outperform the neural approach, especially if there is no pre-trained model supporting that particular language~\cite{winata2022nusax}. Moreover, with the limited computational access in many regions such as Indonesia, classical machine learning remains a popular choice for researchers and industry~\cite{nityasya2020costs,aji2022one}. Hence, we utilize this approach for \datasetname{}.

For NLU tasks, we employ three classical machine learning methods as our baselines, namely (1) Naive Bayes~\cite{zhang2004optimality}, (2) Logistic Regression~\cite{Cramer2003}, and (3) SVM~\cite{scholkopf1995svm}. For NLG tasks, we harness three methods to do benchmarking on machine translation tasks; (1) direct copy from the source language, in this case, Indonesian; (2) word lexical substitution via bilingual Panlex lexicons; and (3) phrase-based statistical machine translation (PBSMT)~\cite{koehn-etal-2003-statistical}. We employ the PBSMT method from Moses toolkit \cite{koehn2007moses}.

\paragraph{Massively Multilingual LMs}

Fine-tuning LMs for downstream tasks has become a popular method in NLP. It enables LMs to learn with a limited dataset and perform better compared to training neural models from scratch~\cite{devlin2019bert, wilie2020indonlu,gehrmann-etal-2022-gemv2}.

Moreover, recent work has shown that a fine-tuned model for a specific task can outperform general-purpose, larger language models~\cite{bang2023multitask,asai2023buffet,zhang2023multilingual}. We investigate the performances of both large pre-trained multilingual and Indonesian monolingual baseline models on low-resource languages used in this work. We follow the hyperparameter settings in~\cite{winata2022nusax}. Details are in Appendix~\ref{app:hyperparameters}.

For NLU tasks, we experiment with emotion recognition, sentiment analysis, topic modeling, and rhetoric mode classification. The models used are: (1) mBERT~\cite{devlin2019bert}; (2) IndoNLU~\cite{wilie2020indonlu}; (3) IndoLEM~\cite{koto2020indolem}; and (4) XLM-R~\cite{conneau-etal-2020-unsupervised}.
For NLG tasks, we experiment on machine translation using the following baselines: (1) IndoGPT~\cite{cahyawijaya2021indonlg}; (2) IndoBART~\cite{cahyawijaya2021indonlg}; (3) mBART~\cite{liu2020mbart}; and (4) mT5~\cite{xue-etal-2021-mt5}.

\begin{figure}
    \centering
    \resizebox{\linewidth}{!}{
    \includegraphics{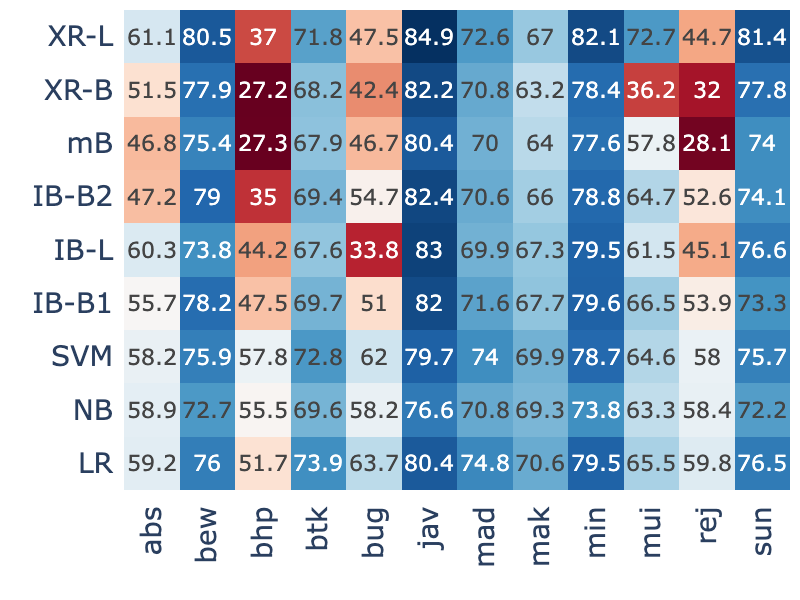}
    }
    \vspace{-1.8em}
    \caption{Per language scores of the classical and fine-tuned baselines. From top to bottom: XLM-R$_\text{LARGE}$, XLM-R$_\text{BASE}$, mBERT, IndoLEM IndoBERT$_\text{BASE}$, IndoNLU IndoBERT$_\text{LARGE}$, IndoNLU IndoBERT$_\text{BASE}$, SVM, Naive Bayes, Logistic Regression.}
    \label{fig:nlu-per-lang}
\end{figure}

\paragraph{Zero-Shot LLMs} 

LLMs fine-tuned through diverse instructions show capabilities to generalize across unseen instructions~\cite{wei2021finetuned, sanh2021multitask,ouyang2022training,yong2023prompting}. Moreover, these models are shown to be able to generalize across different languages, assuming the base model is multilingual~\cite{muennighoff2022crosslingual,adilazuarda2023the,zhang2023multilingual}. Therefore, to assess the zero-shot capabilities of LLMs over our datasets, we benchmark BLOOMZ and mT0~\cite{muennighoff2022crosslingual}, both of which are multilingual LLMs that have been fine-tuned with downstream task instructions. We explore the model from 300M up to $\sim$13B parameters. For NLU, the class output is determined by selecting the most probable label generated after the prompt. For NLG, we generate the translation by using prompts. The prompts used in this experiment can be found in Appendix~\ref{app:prompts}. 

\section{Benchmark Results and Discussion}
\label{sec:results}

We present the results of our NLU and NLG experiments in Table~\ref{tab:nlu_results} and Table~\ref{tab:nlg_results}, respectively.
While the classical baselines have never learned any prior language representations, they perform competitively to the fine-tuning baselines---the fine-tuned Indonesian monolingual models (i.e., IndoBERT, IndoBART, and IndoGPT) and the fine-tuned multilingual models (i.e., mBERT, XLM-R, mBART-50, and mT5)---on both NLU and NLG benchmarks. Furthermore, based on the per language breakdown shown in Figure~\ref{fig:nlu-per-lang}, except for the languages observed during the pre-training, i.e., Javanese (jav) and Sundanese (sun), both Indonesian and multilingual LMs fail to outperform the classical machine learning approaches on most languages and only able to outperform on languages that are closely related to Indonesian (see Appendix~\ref{app:lang_study}), i.e., Betawi (bew) and Minangkabau (min). These facts demonstrate that most extremely low-resource languages in NusaParagraph and NusaTranslation are beyond the scope of the knowledge transfer from Indonesian and multilingual pre-training due to their distinct linguistic characteristics.


Secondly, the LLMs used in this study: BLOOMZ and mT0, consistently and significantly underperform the fine-tuned and classical baselines, e.g., up to $\sim$56\% gap on emotion recognition and $\sim$47\% on topic modeling in NusaParagraph, as well as $\sim$17.5 SacreBLEU on machine translation. Despite their ability to generalize to unseen tasks~\cite{muennighoff2022crosslingual}, LLMs do not generalize well to unseen languages, which indicates a challenge on knowledge transferability between languages, especially for underrepresented and extremely low-resource language, and underlines the need for more language-inclusive LLMs.

\section{Conclusion}
In this work, we compare the effectiveness of corpus collection methods for underrepresented and extremely low-resource languages. From our thorough study, we conclude that, although online scraping is effective for high-resource languages, it is not ideal for many extremely low-resource languages. Other approaches such as sentence translation and paragraph writing can be a better alternative for collecting data in extremely low-resource languages because they produce a better corpus with higher lexical diversity and cultural relevance. Furthermore, to measure the capability of existing LLMs to process underrepresented and extremely low-resource languages, we propose the \datasetname{} benchmark, which covers 12 Indonesian local languages. Based on the benchmarking results, we demonstrate that both existing zero-shot prompting LLMs and fine-tuned pre-trained LMs fail to outperform the classical baselines, suggesting that LMs cannot generalize to these extremely low-resource languages as most of the extremely low-resource languages under study are distinct from other previously learned languages. Our empirical experiments emphasize the need to extend the language coverage of the models.

\section*{Limitations}

\subsection{Languages for Comparison of Corpus Collection Methods}

We explore only 5 Indonesian local languages to compare the effectiveness of different corpus collection methods due to the difficulty of finding eligible annotator candidates for the other languages. We hope future work can explore the generalization of our analysis in broader languages, especially for other underrepresented and extremely low-resource languages in different language families.

\subsection{Buginese Data for NusaTranslation}

We do not have Buginese data in our NusaTranslation corpus, this is due to the difficulty of finding eligible annotator candidates for Buginese. In fact, during our course of dataset construction, we only found one eligible annotator candidate who would like to participate in our study. 

\subsection{Few-Shot LLM Prompting}

Few-shot in-context learning has been shown to be able to improve the performance of zero-shot prompting~\cite{brown2020gpt3,sanh2022t0,wei2022flan,chung2022flan-t5}. However, few-shot in-context learning incurs a high computational cost and, due to a limited computational budget, we only explore zero-shot LLM prompting and we leave the exploration on few-shot in-context learning for future works.

\section*{Acknowledgments}
We are grateful to Shreyashee Sinha for the feedback on a draft of this manuscript. This work has been partially funded by PT. Darta Media Indonesia (Kaskus.co.id) (001/DMI//KS/IV/2022); PhD Fellowship Award, the Hong Kong University of Science and Technology; and PF20-43679 Hong Kong PhD Fellowship Scheme, Research Grant Council, Hong Kong.

\section*{Ethical Consideration}

Our work highlights the importance of democratizing access to Natural Language Processing (NLP) technology for underrepresented and extremely low-resource languages.
During our study, we are well aware of the ethical responsibility associated with language research and the potential impact it can have on communities. 
Our study prioritizes inclusivity, cultural relevance, and fairness. Within this work, the annotators are properly rewarded above the national average minimum wage in Indonesia. We have obtained informed consent from all annotators and adhered to data protection and privacy regulations for releasing the corpus and benchmark. Throughout our research process, we have made conscious efforts to engage with the language communities, involve local experts, and respect their linguistic and cultural nuances. 
Our ultimate goal is to promote linguistic diversity and contribute to a more inclusive NLP landscape providing social good through our work to society, especially in the field of NLP. We encourage further collaboration and engagement with underrepresented language communities to ensure that their voices are heard and their needs are addressed in future language technology development.


\bibliography{custom}
\bibliographystyle{acl_natbib}

\newpage

\appendix















\section{Indonesian Local Languages in the Bible Corpus}
\label{app:bible-corpus}

We list out all the Indonesian local languages that are covered in the Bible corpus in Table~\ref{tab:lang-bible}.

\begin{table}[!ht]
    \centering
    \resizebox{0.8\linewidth}{!}{
        \begin{tabular}{ccc}
        \toprule
        \textbf{Language} & \textbf{ISO code} & \textbf{\#Speakers} \\
        \midrule
		 Balantak  & blz & 33,000 \\
		 Nggem  & nbq & 4,400 \\
		 Alune  & alp & 21,300 \\
		 Bambam  & ptu & 40,000 \\
		 Yawa  & yva & 10,000 \\
		 Dhao  & nfa & 5,000 \\
		 Helong  & heg & 14,000 \\
		 Kupang Malay & mkn & 200,000 \\
		 Mamasa  & mqj & 100,000 \\
		 Luang  & lex & 18,000 \\
		 Ambai  & amk & 10,000 \\
		 Sabu  & hvn & 8,000 \\
		 Amarasi  & aaz & 70,000 \\
		 Kisar  & kje & 20,000 \\
        \bottomrule
        \end{tabular}
    }
    \caption{Description of Indonesian local languages covered in the Bible corpus.}
    \label{tab:lang-bible}
\end{table}

\section{Pre-Annotation Procedure}
\label{app:pre-annotation}

\subsection{Annotator Recruitment}

Our recruitment process involves multiple steps. Firstly, we conduct a strict selection process to filter out applicants. Subsequently, we proceed with knowledge transfer sessions for the selected annotators. The primary objective of our recruitment process is to identify and engage proficient annotators with expertise in relevant local languages.

\paragraph{Qualification} In developing data for a local language, a competent and experienced team in the required local language is certainly needed. Annotators play a crucial role in compiling high-quality local language data. Therefore, strict qualifications are required for the candidate annotators who will be recruited. The qualifications include educational background and experience related to language. Annotator candidates must have good knowledge of the language and the sentence structure of the local language they are proficient in. 

\paragraph{Recruitment Process} The recruitment process starts with an assessment test comprising three questions for each task. This test is designed to provide an overview of the candidate's abilities in sentence translation and paragraph creation in the relevant local language for future tasks. During this stage, the priority in candidate selection is based on the assessment test results, followed by employment status and educational background. From this process, we gathered a total of 892 annotator candidates from different languages. There are 29 candidates for Madurese, 141 for Minangkabau, 319 for Javanese, 217 for Sundanese, 28 for Betawi, 52 for Batak, 45 for Makassarese (including Bugisnese), and 65 for other languages (including Acehnese, Ambonese, Rejang Lebong, Sumbawanese, Papuan, Balinese, Bimanese, Cirebonese, Dayak, Leti, Lombokese, Pontianak Malay, Palembangese, and Tolaki). 

Out of a total of 892 applicants, only 127 candidates ($\sim$14\%) were eligible to participate in the annotation process, and among which, only 83 ($\sim$65\%) candidates expressed their willingness to proceed. Some of the annotators withdraw during the course of the annotations which further increases the complexity of the recruitment process. With this obstacle, the recruitment process faces complex challenges. Finding speakers of certain local languages can be difficult, making the recruitment process long and ongoing throughout the annotation process. 

\paragraph{Knowledge Transfer}

All the selected annotators will join groups and receive explanations regarding this project through knowledge transfer and overview meetings before starting their work. The information provided covers various aspects related to project management and the annotation process in detail. Annotators will gain a clear understanding of the methods and guidelines to be followed in performing these annotations. With this explanation, it is expected that the annotator will have a comprehensive understanding of their responsibilities in this work and a detailed understanding of the task. This will assist them in carrying out their task effectively and producing high-quality output.

\section{Sentence Translation Procedure}
\label{app:translation-proc}
Human translation is carried out by determining the boundaries of the rules in the translation process. We instructed the annotators to retain the meaning of the text and to keep entities, such as persons, organizations, locations, and time with no target language translation the same. Specifically, we instructed them to: (1) maintain the sentence's sentiment polarity; (2) preserve entities; and (3) maintain the complete information content of the original text.

Besides, we asked the annotators to maintain the typography. Most sentences from the original dataset are written in an informal tone, with non-standard spelling, e.g., elongated vowels and punctuation. When the sentence is translated into the target language, direct translation can sound unnatural. For example, translating the Indonesian word \ex{kangeeeen} (originally \ex{kangen}; en: \ex{miss}) to \ex{taragaaaak} (originally \ex{taragak}) in Minangkabau may sound unnatural. Similarly, the original sentence may also contain typos. Due to the difficulty of accurately assessing the typographical consistency of translations, we removed this as a criterion.

The translation annotation phase is planned to last for approximately 2–6 weeks depending on the number of annotators involved in one language group. Each annotator gets around 1,000--3,000 sentences (with the same reasons as the previous explanation). Each annotator is required to complete a translation of 500 sentences per week. However, there were issues of commitment to achieving weekly targets and availability of annotators, extending the annotation process to 9 weeks.

This translation method achieved a total of 72,444 sentences. The details are: 1,579 sentences for Bima language, 1,574 sentences each for Palembang, Rejang Lebong, and Ambon languages, 9,449 sentences each for Madura language, Minangkabau language, Batak language, Betawi language, Javanese, Sundanese, and Makassar language.

\section{List of Topics for Paragraph Writing}
\label{app:topic-selection}

Here we provide the list of topics and subtopics for the paragraph writing data collection. Each topic consists of 10 subtopics unless stated otherwise.

\begin{enumerate}
    \item \textbf{Food \& Beverages.} News about food; Restaurant reviews and fast food; Recipes; Favorite food and disliked food; Favorite drink and disliked drink; Favorite snacks and disliked snacks; Professions related to cuisine; Cooking utensils and kitchens; Vegetable and fruit; Seafood.
    \item \textbf{Leisure.} Nature tourist attractions; Hidden tourist attractions (hidden gems); Popular tourist spots; Hotel and lodging; Transportation for tourism; The most memorable vacation experience; Vacation activities with friends/family; Activity in leisure time; Watching movies; Hobby.
    \item \textbf{Religion.} Routine religious activities; Religious figures; The story in the scriptures; Daily stories related to religion; The last religious sermon that you heard; Religious Holidays; Religious scriptures; The most memorable religious teachings; Religious ceremony; House of worship.
    \item \textbf{Culture \& Heritage.} Traditional event; Traditional houses; Folk songs; Folklores from local regions; Traditional weapon; Special souvenirs from local regions; Traditional musical instruments; Regional/traditional dance; Traditional figure; Regional specialty cuisine.
    \item \textbf{Sports.} Favorite exercise as a child; Easy light exercise; Favorite athlete(s); Sports equipment; Sports at school (extracurricular activity); Sports match; Extreme sports; Unexpected events during sports; Benefits of sports; Watching a sports match.
    \item \textbf{Technology.} Favorite video game; Handphone; Laptop; Television and radio; Washing machine; Camera; Other electronic appliances; Robot; News about technology; Latest technology.
    \item \textbf{Business (5 topics).} Businessman; Work at office; Ideas to sell stuff from home; Tips while losing money; Online selling from the internet.
    \item \textbf{Science.} Animal; Plant; Energy sources; Discoveries; Known figures or researchers; Environmental problems; Diseases and other disorders; Planet and the solar system; Favorite subjects at school; Scientific experiments conducted at school.
    \item \textbf{History.} The history of the house that is inhabited; The history of public facilities in the neighborhood; The history of the city where you grew up; History of Indonesian independence; National and local heroes.
    \item \textbf{Politics.} Favorite and disliked political parties; Favorite and disliked political figures; Known political teachings; Election stories; Rules, laws, known regulations.
    \item \textbf{Emotion: Happy.} Happy to be accepted at college; Happy to pass the exam; Happy to buy goods after saving; Happy because of winning the lottery; Happy to find the item you are looking for; The most fun experience in life; Happy for winning an award; Happy to meet favorite idol(s); Marriage; Happy because of childbirth.
    \item \textbf{Emotion: Sad.} Sad due to layoff; Sad due to the death of family members; Failure in life; Childhood trauma; Sad due to failing university entrance exam; Disappointing beloved people; The saddest experience in life; Deepest regret for life; Sad because getting separated from friends/parents' divorce; Sad because of loneliness.
    \item \textbf{Everyday Life.} Living with a partner/significant other; Preparing an emergency fund; Living with neighbors; How to survive on the mountain; Friendship; Preparing for natural disasters; How to get rid of stress; Farming; Lifestyle in the village; How to ride public transportation.
    \item \textbf{Emotion: Surprise.} Shocked to see a ghost; Surprised to get a present; Shocked to hear an illness diagnosis; Shocked to because of a prank; Surprised to miss the plane/bus; Surprised to accidentally meet old friends; Surprised to win the lottery; Shocked by positive pregnancy test; Surprised to get a birthday surprise; Surprised to find treasure.
    \item \textbf{Emotion: Angry.} Angry because someone skipped the queue; Angry because of an insult about physique; Angry to get cheated on; Angry to get hit; Angry because of a traffic jam; Angry without reason; Angry because of unfair treatment; Angry because of discrimination; Angry because other people take our work; Angry at an officer.
    \item \textbf{Emotion: Fear (20 subtopics).} Scared because of stage fright; Afraid to be at a gunpoint/knifepoint; Fear that evolves to trauma (phobia); Afraid and at the death's door; Scared of seeing ghosts; Afraid of being bullied; Afraid due to being threatened; Fear of drowning; Fear of surgery; Fear of being scolded; Fear of needles; Fear when committing sin/mistakes; Afraid to be a witness to an accident; Afraid when natural disasters occur; Afraid when being chased by animals; Afraid when being stalked; Fear of being alone; Panic attack; Afraid of being in a strange place; Scared when the brake fails.
    \item \textbf{Emotion: Disgusted (5 subtopics).} Disgusted at vomit; Disgusted at a dirty toilet; Disgusted with animal waste; Disgusting fishy smell; Disgusted at a house full of insects; Disgusted by other people's saliva; Disgusted at a pile of trash; Disgusted at rotten food; Disgusted at dirty food; Other disgusting experiences/stories.
    \item \textbf{Emotion: Shame (5 subtopics).} Embarrassed because of torn pants; Embarrassed of farting in public places; Embarrassed because of accidentally getting into the wrong toilet; Embarrassed to be laughed at by classmate(s); Embarrassed to misidentify someone; Embarrassed because of accidentally wearing the same clothes as strangers; Other embarrassing experiences; Embarrassed to get caught red-handed; Embarrassed to be stinky; Embarrassed to be in debt.
    \item \textbf{Stance: Support/Neutral/Contradict (20 subtopics).} Abortion; Atheism; 1 week = 4 working days + 3 holidays compared to 1 week = 5 working days + 2 holidays; The elimination of national examination; Celibacy; Liberalism; Socialism; Communism; Body positivity; LGBTQ+ in social life; Cloning; The existence of spirits (ghosts/demons/etc.); Reincarnation; The need for college; Culture preservation; Panda conservation; The need for shaving leg hair; Death penalty; Friendship between men and women without being more than friends; The legalization of assisted suicide.
\end{enumerate}

\begin{figure*}[t]
     \centering
     \begin{subfigure}[b]{0.3\textwidth}
         \centering
         \includegraphics[width=\textwidth]{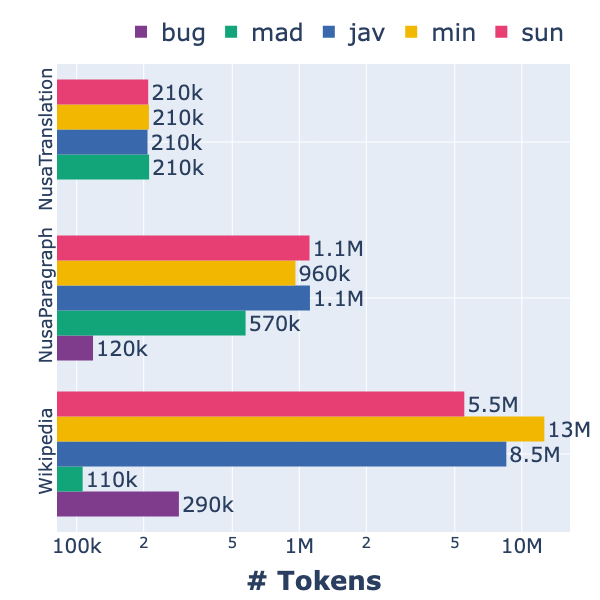}
     \end{subfigure}
     \begin{subfigure}[b]{0.3\textwidth}
         \centering
         \includegraphics[width=\textwidth]{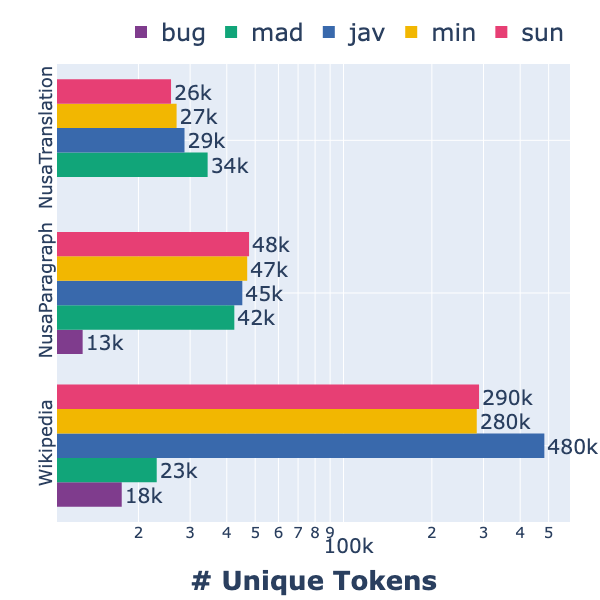}
     \end{subfigure}
     \begin{subfigure}[b]{0.3\textwidth}
         \centering
         \includegraphics[width=\textwidth]{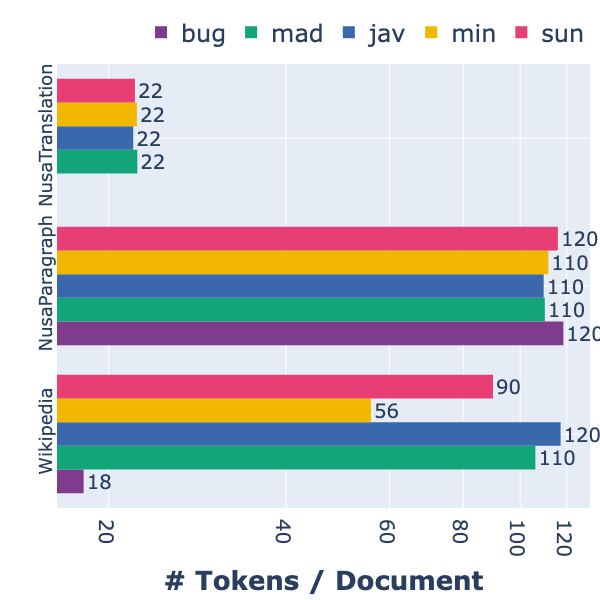}
     \end{subfigure}
     \caption{\textbf{(left)} \#Tokens, \textbf{(center)} \#Unique Tokens, and \textbf{(right)} \#Tokens/Document statistics of the corpus collection methods under study: NusaTranslation, NusaParagraph, and Wikipedia. }
     \label{fig:general-statistics}
\end{figure*}

\section{Paragraph Writing Procedure}
\label{app:paragraph-writing}

For paragraph writing, we initially provide a list of topics before instructing the annotators to write paragraphs. The topics provided vary widely, ranging from simple topics such as food and beverages, entertainment/leisure, and sports, to quite heavy topics such as science, history, politics, and religion. We also provided more specific subtopics from each of the major topics provided. In total, we have 20 main topics with 20 subtopics of each main topic.

The provision of topics (especially, subtopics) aims to facilitate the annotators in the process of writing paragraphs. That way, the annotators only need to write paragraphs by developing ideas from the topics and subtopics that have been given without the need to think about which topic to choose. In addition, the selection of various topics is also expected to enrich vocabulary in the corpus; different topics will obviously bring up the use of different dictionaries.

For conducting the paragraph writing, the annotators are instructed to write short paragraphs with the following criteria: (a) the paragraph consists of a minimum of 100 words, (b) using the targeted local language, (c) the topic is according to the provided topics and subtopics, (d) the type of the paragraphs are narration, description, argumentation, persuasion, and exposition, (e) The content must not defame the name of public entities or contain sensitive and personal information of specific individuals.

The paragraph writing procedure is started (1) after a transfer of learning given to the annotators about the general procedure and knowledge about writing and paragraph types. (2) After that, every annotator is given access to their own worksheet in Google Spreadsheet that already contains all topics and subtopics that they can develop according to the procedure. Every annotator had time around 15 weeks to finish 100–160 paragraphs every week. (3) While annotators already start their paragraph writing process, QC annotators check and validate their work every week (around 2–3 times a week) to the paragraphs that are already done. (4) Lastly, every two weeks, every annotator is gathered in an online meeting to discuss the evaluation of any errors found in their data to prevent mistakes in the future.

Through paragraph writing, we achieved a total of 56395 paragraphs. The details are: 5017 paragraphs for Madura language, 8538 paragraphs for Minangkabau, 10189 paragraphs for Javanese, 9729 paragraphs for Sundanese, 9756 paragraphs for Betawi, 4711 paragraphs for Batak, 5338 paragraphs for Makassar, 1200 paragraphs for Rejang Lebong, 1473 paragraphs for Palembang, 1059 paragraphs for  Bugis language, and 44 paragraphs for Ambon language.

\section{Post Annotation Procedure}
\label{app:data-validation}

\subsection{Manual Validation} For sentence translation, QC annotators check manually through the data to ensure that all words are translated to the target language and not a single word is skipped by the translator. For paragraph writing, QC annotators check the data by skimming through the paragraphs one by one, checking for any apparent typos, and making sure that the annotators are using the local language and not Indonesian. There are some cases where local languages still use Indonesian words, but it should only be below 30\%, while most of the paragraphs must be in the desired local language. To ensure there is no plagiarism,
QC annotators also check by sampling some paragraphs from the data and check whether a similar paragraph is found through a search engine.

\subsection{Automatic Validation}
To further ensure the diversity of the samples, we run an automatic validation to ensure there are no similar paragraphs written by any annotators. Our automatic validation matches two paragraphs by first removing all punctuation marks and then performing string matching using Levensthein distance, and normalizing the distance by dividing with the average length of the two paragraphs. We conduct the process for all the paragraph pairs and we ask the corresponding annotators to revise when the normalized distance of two paragraphs is less than 30\%.

\begin{table*}[!t]
    \centering
    \resizebox{\linewidth}{!}{
        \begin{tabular}{l|c|c|l|c|c|l|c|c|l|c|c}
            \toprule
            \multicolumn{3}{c|}{\textbf{Javanese (jav)}} & \multicolumn{3}{c|}{\textbf{Madurese (mad)}} & \multicolumn{3}{c|}{\textbf{Minangkabau (min)}} & \multicolumn{3}{c}{\textbf{Sundanese (sun)}} \\ 
            \midrule
            \textbf{word} & \textbf{freq}. & \textbf{prop. (\%)} & \textbf{word} & \textbf{freq}. & \textbf{prop. (\%)} & \textbf{word} & \textbf{freq}. & \textbf{prop. (\%)} & \textbf{word} & \textbf{freq}. & \textbf{prop. (\%)} \\
            \midrule
            ac & 380 & 0.18 & ac & 482 & 0.23 & ac & 473 & 0.22 & ac & 460 & 0.22 \\
            wifi & 278 & 0.13 & wifi & 294 & 0.14 & wifi & 311 & 0.15 & wifi & 290 & 0.14 \\
            airy & 273 & 0.13 & airy & 270 & 0.13 & airy & 278 & 0.13 & airy & 277 & 0.13 \\
            hotel & 243 & 0.12 & tv & 255 & 0.12 & tv & 259 & 0.12 & tv & 259 & 0.12 \\
            tv & 227 & 0.11 & hotel & 242 & 0.11 & hotel & 236 & 0.11 & hotel & 249 & 0.12 \\
            hp & 201 & 0.10 & hp & 202 & 0.10 & hp & 225 & 0.11 & hp & 181 & 0.09 \\
            video & 76 & 0.04 & wc & 99 & 0.05 & mode & 122 & 0.06 & via & 87 & 0.04 \\
            paste & 94 & 0.04 & via & 116 & 0.05 & motor & 61 & 0.03 & & & \\
            via & 89 & 0.04 & shower & 89 & 0.04 & twitter & 52 & 0.02 & & & \\
            video & 77 & 0.04 & wc & 77 & 0.04 & & & & & & \\ 
            room & 58 & 0.03 & & & & & & & & & \\ 
            mode & 57 & 0.03 & & & & & & & & & \\
            \bottomrule
        \end{tabular}
    }
    \caption{Common loan words in NusaTranslation from top-200 overlap with the English lexicon and loan word list.}
    \label{tab:mt-borrow}
\end{table*}

\begin{table*}[!t]
    \centering
    \resizebox{\linewidth}{!}{
        \begin{tabular}{l|c|c|l|c|c|l|c|c|l|c|c|l|c|c}
            \toprule
            \multicolumn{3}{c|}{\textbf{Buginese (bug)}} & \multicolumn{3}{c|}{\textbf{Javanese (jav)}} & \multicolumn{3}{c|}{\textbf{Madurese (mad)}} & \multicolumn{3}{c|}{\textbf{Minangkabau (min)}} & \multicolumn{3}{c}{\textbf{Sundanese (sun)}} \\ 
            \midrule
            \textbf{word} & \textbf{freq}. & \textbf{prop. (\%)} & \textbf{word} & \textbf{freq}. & \textbf{prop. (\%)} & \textbf{word} & \textbf{freq}. & \textbf{prop. (\%)} & \textbf{word} & \textbf{freq}. & \textbf{prop. (\%)} & \textbf{word} & \textbf{freq}. & \textbf{prop. (\%)} \\
            \midrule
            media & 683 & 0.58 & online & 448 & 0.04 & wedding & 165 & 0.03 & film & 333 & 0.03 & film & 548 & 0.05 \\
            hp & 535 & 0.45 & media & 432 & 0.04 & laptop & 83 & 0.01 & motor & 253 & 0.03 & duo & 543 & 0.05 \\
            tv & 396 & 0.33 & laptop & 424 & 0.04 & online & 82 & 0.01 & tv & 239 & 0.02 & hp & 410 & 0.04 \\
            laptop & 377 & 0.32 & instagram & 337 & 0.04 & wc & 69 & 0.01 & media & 182 & 0.02 & media & 407 & 0.04 \\
            hotel & 304 & 0.26 & tv & 291 & 0.04 & bully & 42 & 0.01 & laptop & 172 & 0.02 & tv & 333 & 0.03 \\
            video & 298 & 0.25 & robot & 275 & 0.04 & media & 35 & 0.01 & robot & 155 & 0.02 & laptop & 276 & 0.02 \\
            online & 270 & 0.23 &  &  &  &  &  &  &  &  &  & video & 232 & 0.02 \\
            internet & 258 & 0.22 &  &  &  &  &  &  &  &  &  &  &  & \\
            \bottomrule
        \end{tabular}
    }
    \caption{Common loan words in NusaParagraph from top-200 overlap with the English lexicon and loan word list.}
    \label{tab:para-borrow}
\end{table*}

\section{Token Statistics of the Corpora Under Study}
\label{app:token-statistics}

We provide the token statistics of Wikipedia, NusaTranslation, and NusaParagraph in Figure~\ref{fig:general-statistics}. Especially in Buginese (bug), the document length and the number of unique tokens in Wikipedia are rather low, indicating that there is a lot of boilerplate text in the Buginese Wikipedia data.

\section{List of Loan Words in Indonesian Local Languages}
\label{app:borrowed-words}

We present the list of manually curated loan words with their frequency and proportion in the corresponding corpus for each language in Table~\ref{tab:mt-borrow}, Table~\ref{tab:para-borrow}, and Table~\ref{tab:wiki-borrow}, for NusaTranslation, NusaParagraph, and Wikipedia, respectively.

\begin{table*}[!t]
    \centering
    \resizebox{\linewidth}{!}{
        \begin{tabular}{l|c|c|l|c|c|l|c|c|l|c|c|l|c|c}
            \toprule
            \multicolumn{3}{c|}{\textbf{Buginese (bug)}} & \multicolumn{3}{c|}{\textbf{Javanese (jav)}} & \multicolumn{3}{c|}{\textbf{Madurese (mad)}} & \multicolumn{3}{c|}{\textbf{Minangkabau (min)}} & \multicolumn{3}{c}{\textbf{Sundanese (sun)}} \\ 
            \midrule
            \textbf{word} & \textbf{freq}. & \textbf{prop. (\%)} & \textbf{word} & \textbf{freq}. & \textbf{prop. (\%)} & \textbf{word} & \textbf{freq}. & \textbf{prop. (\%)} & \textbf{word} & \textbf{freq}. & \textbf{prop. (\%)} & \textbf{word} & \textbf{freq}. & \textbf{prop. (\%)} \\
            \midrule
            kaisne & 2450 & 0.85 & the & 22125 & 0.26 & of & 554 & 0.52 & asteroid & 251354 & 1.99 & the & 51608 & 0.94 \\ 
            somme & 2323 & 0.81 & of & 18454 & 0.22 & planet & 301 & 0.28 & ordo & 159089 & 1.26 & apollo & 23442 & 0.43 \\ 
            eure & 2027 & 0.70 & and & 8649 & 0.10 & and & 257 & 0.24 & filum & 151064 & 1.20 & of & 23181 & 0.42 \\ 
            manche & 1805 & 0.63 & a & 7256 & 0.09 & gregorian & 225 & 0.21 & animalia & 148731 & 1.18 & amor & 19682 & 0.36 \\ 
            hautegaronne & 1770 & 0.61 & data & 3809 & 0.04 & wikipedia & 207 & 0.19 & kingdom & 148008 & 1.17 & international & 15804 & 0.29 \\ 
            dordogne & 1673 & 0.58 & web & 3629 & 0.04 & influenza & 203 & 0.19 & arthropoda & 147996 & 1.17 & planet & 15321 & 0.28 \\ 
            hautesaône & 1637 & 0.57 & new & 3482 & 0.04 &  &  &  & insecta & 146159 & 1.16 & center & 15260 & 0.28 \\ 
            gironde & 1628 & 0.57 & film & 6100 & 0.07 &  &  &  & cerambycidae & 101158 & 0.80 & union & 15247 & 0.28 \\ 
            vosges & 1547 & 0.54 & for & 2731 & 0.03 &  &  &  & diptera & 94388 & 0.75 & orbit & 15242 & 0.28 \\ 
            hautespyrénées & 1424 & 0.49 & to & 2651 & 0.03 &  &  &  & nebula & 88656 & 0.70 & minor & 15054 & 0.27 \\ 
            gers & 1391 & 0.48 & tv & 2421 & 0.03 &  &  &  & planet & 66110 & 0.52 & asteroid & 15015 & 0.27 \\ 
            ardennes & 1391 & 0.48 & no & 2267 & 0.03 &  &  &  & database & 58923 & 0.47 & astronomical & 14970 & 0.27 \\ 
            yonne & 1363 & 0.47 & as & 2262 & 0.03 &  &  &  & larva & 52188 & 0.41 & primordial & 14931 & 0.27 \\ 
            hautemarne & 1302 & 0.45 & family & 2260 & 0.03 &  &  &  & coleoptera & 51860 & 0.41 & and & 14887 & 0.27 \\ 
            ain & 1259 & 0.44 & john & 2063 & 0.02 &  &  &  & planetesimal & 50875 & 0.40 & for & 4663 & 0.08 \\ 
            eureetloir & 1211 & 0.42 & isbn & 2058 & 0.02 &  &  &  & primordial & 46516 & 0.37 & by & 3536 & 0.06 \\ 
            hautrhin & 1133 & 0.39 & world & 2046 & 0.02 &  &  &  & titan & 33741 & 0.27 & that & 3303 & 0.06 \\ 
            drôme & 1109 & 0.39 & university & 2019 & 0.02 &  &  &  & the & 33386 & 0.26 & are & 3235 & 0.06 \\ 
            gard & 1063 & 0.37 &  &  &  &  &  &  & orbit & 23121 & 0.18 & with & 2859 & 0.05 \\ 
            ardèche & 1020 & 0.35 &  &  &  &  &  &  & limoniidae & 21274 & 0.17 & be & 2636 & 0.05 \\ 
            ariège & 998 & 0.35 &  &  &  &  &  &  & tachinidae & 20520 & 0.16 & or & 2483 & 0.05 \\ 
            allier & 963 & 0.33 &  &  &  &  &  &  & linear & 19213 & 0.15 & as & 5071 & 0.09 \\ 
            deuxsèvres & 917 & 0.32 &  &  &  &  &  &  & socorro & 19196 & 0.15 & this & 2027 & 0.04 \\ 
            aube & 868 & 0.30 &  &  &  &  &  &  & antena & 19017 & 0.15 & at & 2002 & 0.04 \\ 
            corrèze & 861 & 0.30 &  &  &  &  &  &  & jpl & 18739 & 0.15 & gregorian & 1728 & 0.03 \\ 
            finistère & 850 & 0.30 &  &  &  &  &  &  & semi & 18407 & 0.15 & which & 1722 & 0.03 \\ 
            picardy & 816 & 0.28 &  &  &  &  &  &  & minor & 18392 & 0.15 & new & 1721 & 0.03 \\ 
            yvelines & 787 & 0.27 &  &  &  &  &  &  & porifera & 18208 & 0.14 & was & 1550 & 0.03 \\ 
            hauteloire & 782 & 0.27 &  &  &  &  &  &  & browser & 17753 & 0.14 & ubar & 1523 & 0.03 \\ 
            maineetloire & 603 & 0.21 &  &  &  &  &  &  & genus & 17721 & 0.14 & not & 1279 & 0.02 \\ 
            alpesdehauteprovence & 602 & 0.21 &  &  &  &  &  &  & international & 17509 & 0.14 &  &  &  \\ 
            vienne & 566 & 0.20 &  &  &  &  &  &  & center & 17313 & 0.14 &  &  &  \\ 
            hautesalpes & 533 & 0.19 &  &  &  &  &  &  & union & 17232 & 0.14 &  &  &  \\ 
            alpesmaritimes & 491 & 0.17 &  &  &  &  &  &  & astronomical & 17184 & 0.14 &  &  &  \\ 
            hautevienne & 403 & 0.14 &  &  &  &  &  &  & of & 16642 & 0.13 &  &  &  \\ 
            essonne & 395 & 0.14 &  &  &  &  &  &  & muscidae & 15810 & 0.13 &  &  &  \\ 
            pyrénéesorientales & 185 & 0.06 &  &  &  &  &  &  & ado & 15738 & 0.12 &  &  &  \\ 
            communauté & 163 & 0.06 &  &  &  &  &  &  & asilidae & 14571 & 0.12 &  &  &  \\ 
            loiretcher & 142 & 0.05 &  &  &  &  &  &  & apollo & 14350 & 0.11 &  &  &  \\ 
            guadeloupe & 99 & 0.03 &  &  &  &  &  &  & demospongiae & 13249 & 0.10 &  &  &  \\ 
            communes & 55 & 0.02 &  &  &  &  &  &  & history & 13053 & 0.10 &  &  &  \\ 
            the & 33 & 0.01 &  &  &  &  &  &  & cecidomyiidae & 13050 & 0.10 &  &  &  \\ 
            of & 32 & 0.01 &  &  &  &  &  &  & american & 12950 & 0.10 &  &  &  \\ 
            community & 15 & 0.01 &  &  &  &  &  &  & spider & 12833 & 0.10 &  &  &  \\ 
            singapore & 14 & 0.00 &  &  &  &  &  &  & catalog & 12832 & 0.10 &  &  &  \\ 
            from & 12 & 0.00 &  &  &  &  &  &  & araneae & 12814 & 0.10 &  &  &  \\ 
            language & 12 & 0.00 &  &  &  &  &  &  & ceratopogonidae & 11324 & 0.09 &  &  &  \\ 
            wedding & 11 & 0.00 &  &  &  &  &  &  & bombyliidae & 10934 & 0.09 &  &  &  \\ 
            proton & 10 & 0.00 &  &  &  &  &  &  & salticidae & 10431 & 0.08 &  &  &  \\ 
            canton & 10 & 0.00 &  &  &  &  &  &  & parasit & 10294 & 0.08 &  &  & \\            
            \bottomrule
        \end{tabular}
    }
    \caption{Common loan words in NusaParagraph from top-200 overlap with the English lexicon and loan word list. We only show the top 50 words for Buginese (bug) and Minangkabau (min). In total, the top 200 overlapping Buginese (bug) and Minangkabau (min) data with the English lexicon contain 54 and 90 loan words, respectively.}
    \label{tab:wiki-borrow}
\end{table*}

\section{List of Common Local Words In Indonesian Local Languages}
\label{app:common-word-ratio}

We present the list of manually curated common local words with their frequency and proportion in the corresponding corpus for each language in Table~\ref{tab:mt-common}, Table~\ref{tab:para-common}, and Table~\ref{tab:wiki-common} for NusaTranslation, and NusaParagraph, Wikipedia, respectively.

\begin{table*}[!t]
    \centering
    \resizebox{\linewidth}{!}{
    \begin{tabular}{l|l|c|c|c|c|c|c|c|c}
        \toprule
        \multirow{2}{*}{\textbf{Topic}} & \multirow{2}{*}{\textbf{Word}} & \multicolumn{2}{c|}{\textbf{Javanese (jav)}} & \multicolumn{2}{c|}{\textbf{Madurese (mad)}} & \multicolumn{2}{c|}{\textbf{Minangkabau (min)}}  & \multicolumn{2}{c}{\textbf{Sundanese (sun)}} \\
        \cmidrule{3-8}
         &  & \textbf{freq} & \textbf{prop (\%)} & \textbf{freq} & \textbf{prop (\%)} & \textbf{freq} & \textbf{prop (\%)} & \textbf{freq} & \textbf{prop (\%)} \\
         \midrule
        \multirow{5}{*}{food} & indomie & 9 & 0.0076 & 7 & 0.0006 & 11 & 0.0019 & 11 & 0.0011 \\
         & rendang & 3 & 0.0025 & 3 & 0.0003 & 0 & 0.0000 & 3 & 0.0003 \\
         & tempe & 3 & 0.0025 & 2 & 0.0002 & 2 & 0.0003 & 3 & 0.0003 \\
         & gule & 0 & 0.0000 & 0 & 0.0000 & 0 & 0.0000 & 0 & 0.0000 \\
         & sate & 3 & 0.0025 & 3 & 0.0003 & 3 & 0.0005 & 3 & 0.0003 \\ \midrule
        \multirow{3}{*}{transportation} & angkot & 6 & 0.0051 & 4 & 0.0004 & 6 & 0.0010 & 5 & 0.0005 \\
         & ojol & 5 & 0.0042 & 4 & 0.0004 & 6 & 0.0010 & 5 & 0.0005 \\
         & gojek & 15 & 0.0127 & 13 & 0.0012 & 15 & 0.0026 & 14 & 0.0015 \\ \midrule
        \multirow{4}{*}{religion} & doa & 1 & 0.0008 & 12 & 0.0011 & 47 & 0.0082 & 36 & 0.0037 \\
         & gaib & 3 & 0.0025 & 2 & 0.0002 & 1 & 0.0002 & 2 & 0.0002 \\
         & alhamdulilah & 2 & 0.0017 & 4 & 0.0004 & 2 & 0.0003 & 14 & 0.0015 \\
         & insyaallah & 4 & 0.0034 & 4 & 0.0004 & 2 & 0.0003 & 1 & 0.0001 \\ \midrule
        \multirow{3}{*}{adjective} & bule & 9 & 0.0076 & 8 & 0.0007 & 16 & 0.0028 & 15 & 0.0016 \\
         & santun & 2 & 0.0017 & 0 & 0.0000 & 2 & 0.0003 & 3 & 0.0003 \\
         & alay & 9 & 0.0076 & 9 & 0.0008 & 11 & 0.0019 & 7 & 0.0007 \\
         \midrule
        \multicolumn{2}{c|}{\textbf{Total}} &  74 & \textbf{0.0625} & \textbf{75} & \textbf{0.0059} & \textbf{124} & \textbf{0.0197} & \textbf{122} & \textbf{0.0120} \\
        \bottomrule
        \end{tabular}
    }
    \caption{Common local words with their frequency and proportion in the NusaTranslation corpus.}
    \label{tab:mt-common}
\end{table*}

\begin{table*}[!t]
    \centering
    \resizebox{\linewidth}{!}{
    \begin{tabular}{l|l|c|c|c|c|c|c|c|c|c|c}
        \toprule
        \multirow{2}{*}{\textbf{Topic}} & \multirow{2}{*}{\textbf{Word}} & \multicolumn{2}{c|}{\textbf{Buginese (bug)}} & \multicolumn{2}{c|}{\textbf{Javanese (jav)}} & \multicolumn{2}{c|}{\textbf{Madurese (mad)}} & \multicolumn{2}{c|}{\textbf{Minangkabau (min)}}  & \multicolumn{2}{c}{\textbf{Sundanese (sun)}} \\
        \cmidrule{3-8}
         &  & \textbf{freq} & \textbf{prop (\%)} & \textbf{freq} & \textbf{prop (\%)} & \textbf{freq} & \textbf{prop (\%)} & \textbf{freq} & \textbf{prop (\%)} & \textbf{freq} & \textbf{prop (\%)} \\
         \midrule
        \multirow{5}{*}{food} & indomie & 0 & 0.0000 & 24 & 0.0021 & 9 & 0.0016 & 12 & 0.0012 & 15 & 0.0013 \\
         & rendang & 7 & 0.0059 & 27 & 0.0024 & 9 & 0.0016 & 4 & 0.0004 & 32 & 0.0029 \\
         & tempe & 2 & 0.0017 & 101 & 0.0090 & 45 & 0.0078 & 29 & 0.0030 & 31 & 0.0028 \\
         & gule & 0 & 0.0000 & 8 & 0.0007 & 32 & 0.0056 & 0 & 0.0000 & 2 & 0.0002 \\
         & sate & 7 & 0.0059 & 69 & 0.0062 & 122 & 0.0213 & 112 & 0.0117 & 91 & 0.0082 \\
        \multirow{3}{*}{transportation} & angkot & 8 & 0.0068 & 52 & 0.0047 & 3 & 0.0005 & 57 & 0.0059 & 254 & 0.0228 \\
         & ojol & 0 & 0.0000 & 6 & 0.0005 & 0 & 0.0000 & 0 & 0.0000 & 16 & 0.0014 \\
         & gojek & 12 & 0.0101 & 15 & 0.0013 & 15 & 0.0026 & 4 & 0.0004 & 17 & 0.0015 \\
        \multirow{4}{*}{religion} & doa & 11 & 0.0093 & 5 & 0.0004 & 3 & 0.0005 & 102 & 0.0106 & 82 & 0.0074 \\
         & gaib & 0 & 0.0000 & 46 & 0.0041 & 1 & 0.0002 & 13 & 0.0014 & 18 & 0.0016 \\
         & alhamdulilah & 0 & 0.0000 & 1 & 0.0001 & 6 & 0.0010 & 2 & 0.0002 & 24 & 0.0022 \\
         & insyaallah & 0 & 0.0000 & 1 & 0.0001 & 12 & 0.0021 & 17 & 0.0018 & 8 & 0.0007 \\
        \multirow{3}{*}{adjective} & bule & 1 & 0.0008 & 5 & 0.0004 & 8 & 0.0014 & 1 & 0.0001 & 9 & 0.0008 \\
         & santun & 0 & 0.0000 & 9 & 0.0008 & 5 & 0.0009 & 11 & 0.0011 & 14 & 0.0013 \\
         & alay & 0 & 0.0000 & 2 & 0.0002 & 3 & 0.0005 & 9 & 0.0009 & 1 & 0.0001 \\
         \midrule
        \multicolumn{2}{c|}{\textbf{Total}} & \textbf{48} &\textbf{ 0.0405} & \textbf{371} & \textbf{0.0332} & \textbf{273} & \textbf{0.0471} & \textbf{373} & \textbf{0.0379} & \textbf{614} & \textbf{0.0551} \\
        \bottomrule
        \end{tabular}
    }
    \caption{Common local words with their frequency and proportion in the NusaParagraph corpus.}
    \label{tab:para-common}
\end{table*}

\begin{table*}[!t]
    \centering
    \resizebox{\linewidth}{!}{
    \begin{tabular}{l|l|c|c|c|c|c|c|c|c|c|c}
        \toprule
        \multirow{2}{*}{\textbf{Topic}} & \multirow{2}{*}{\textbf{Word}} & \multicolumn{2}{c|}{\textbf{Buginese (bug)}} & \multicolumn{2}{c|}{\textbf{Javanese (jav)}} & \multicolumn{2}{c|}{\textbf{Madurese (mad)}} & \multicolumn{2}{c|}{\textbf{Minangkabau (min)}}  & \multicolumn{2}{c}{\textbf{Sundanese (sun)}} \\
        \cmidrule{3-8}
         &  & \textbf{freq} & \textbf{prop (\%)} & \textbf{freq} & \textbf{prop (\%)} & \textbf{freq} & \textbf{prop (\%)} & \textbf{freq} & \textbf{prop (\%)} & \textbf{freq} & \textbf{prop (\%)} \\
         \midrule
        \multirow{5}{*}{food} & indomie & 0 & 0.0000 & 36 & 0.0032 & 1 & 0.0002 & 1 & 0.0001 & 51 & 0.0046 \\
         & rendang & 0 & 0.0000 & 36 & 0.0032 & 0 & 0.0000 & 14 & 0.0015 & 24 & 0.0022 \\
         & tempe & 0 & 0.0000 & 123 & 0.0110 & 0 & 0.0000 & 29 & 0.0030 & 11 & 0.0010 \\
         & gule & 0 & 0.0000 & 26 & 0.0023 & 0 & 0.0000 & 2 & 0.0002 & 4 & 0.0004 \\
         & sate & 0 & 0.0000 & 223 & 0.0200 & 2 & 0.0003 & 169 & 0.0176 & 21 & 0.0019 \\ \midrule
        \multirow{3}{*}{transportation} & angkot & 0 & 0.0000 & 14 & 0.0013 & 0 & 0.0000 & 8 & 0.0008 & 56 & 0.0050 \\
         & ojol & 0 & 0.0000 & 0 & 0.0000 & 0 & 0.0000 & 0 & 0.0000 & 0 & 0.0000 \\
         & gojek & 0 & 0.0000 & 18 & 0.0016 & 1 & 0.0002 & 20 & 0.0021 & 0 & 0.0000 \\ \midrule
        \multirow{4}{*}{religion} & doa & 0 & 0.0000 & 88 & 0.0079 & 2 & 0.0003 & 35 & 0.0036 & 65 & 0.0058 \\
         & gaib & 0 & 0.0000 & 90 & 0.0081 & 0 & 0.0000 & 17 & 0.0018 & 70 & 0.0063 \\
         & alhamdulilah & 0 & 0.0000 & 2 & 0.0002 & 0 & 0.0000 & 0 & 0.0000 & 0 & 0.0000 \\
         & insyaallah & 0 & 0.0000 & 2 & 0.0002 & 0 & 0.0000 & 0 & 0.0000 & 0 & 0.0000 \\ \midrule
        \multirow{3}{*}{adjective} & bule & 0 & 0.0000 & 19 & 0.0017 & 0 & 0.0000 & 5 & 0.0005 & 7 & 0.0006 \\
         & santun & 0 & 0.0000 & 39 & 0.0035 & 0 & 0.0000 & 7 & 0.0007 & 14 & 0.0013 \\
         & alay & 0 & 0.0000 & 7 & 0.0006 & 0 & 0.0000 & 0 & 0.0000 & 0 & 0.0000 \\
         \midrule
        \multicolumn{2}{c|}{\textbf{Total}} & \textbf{0} & \textbf{0.0000} & \textbf{723} & \textbf{0.0647} & \textbf{6} & \textbf{0.0010} & \textbf{307} & \textbf{0.0319} & \textbf{323} & \textbf{0.0291} \\
        \bottomrule
        \end{tabular}
    }
    \caption{Common local words with their frequency and proportion in the Wikipedia corpus.}
    \label{tab:wiki-common}
\end{table*}

 \begin{figure}[h]
     \centering
     \resizebox{\linewidth}{!}{
     \includegraphics[width=\textwidth]{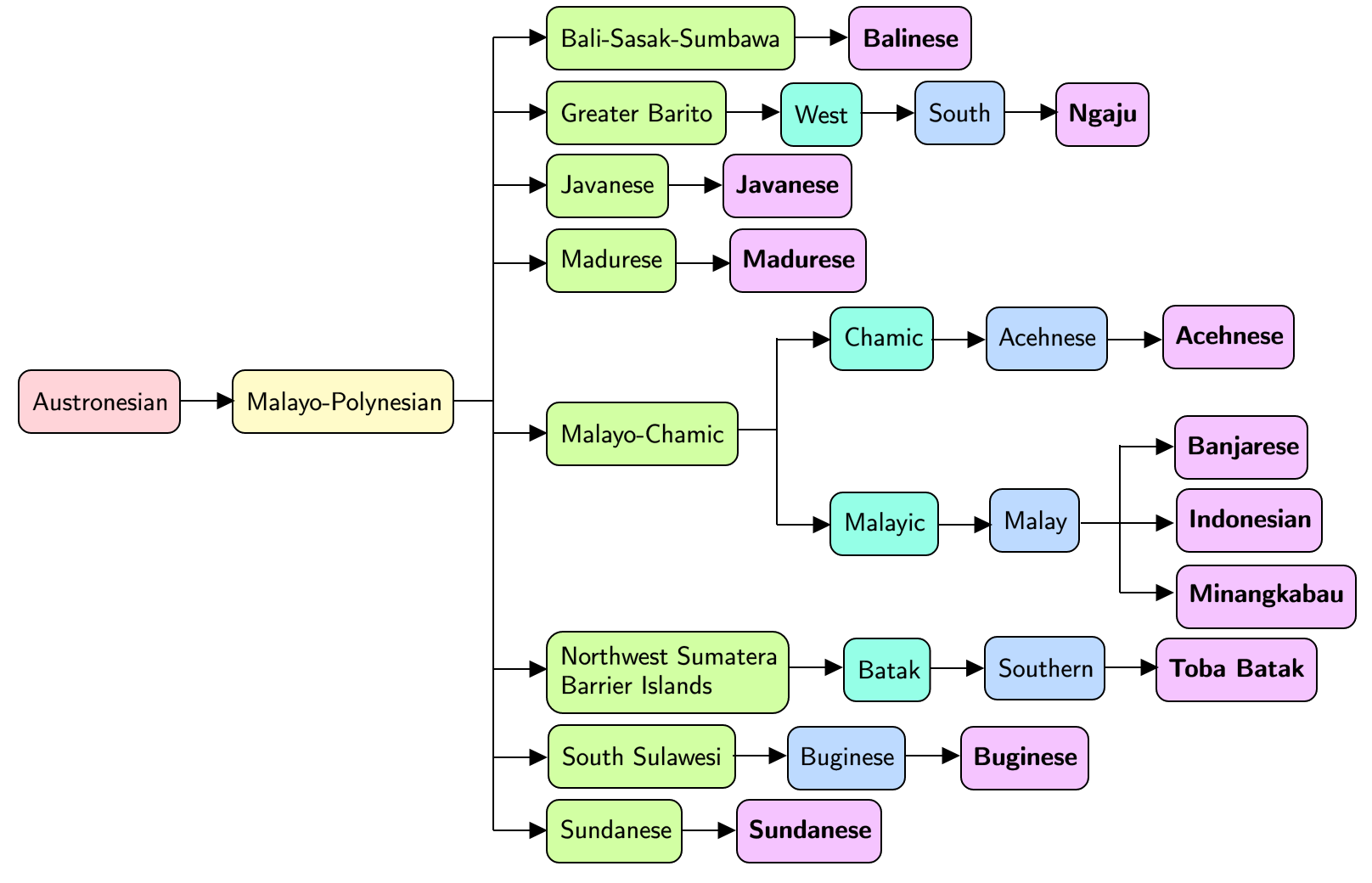}
     }
     \caption{Taxonomy of the languages under study. We show all of the 12 Indonesian local languages under study and the national language of Indonesia, i.e., Indonesian (ind).}
    \label{fig:lang-tree}
 \end{figure}

\section{Languages Under Study}
\label{app:lang_study}

\begin{table}[!t]
     \centering
     \resizebox{0.95\linewidth}{!}{
        \begin{tabular}{lcc}
            \toprule
            \textbf{Language} & \textbf{ISO code} & \textbf{Status} \\
            \midrule
            {Ambonese Malay} & {abs} & wider communication\\
            {Mandailing} & {btm} & threatened\\
            {Batak Toba} & {bbc} & threatened\\
            
            {Betawi} & {bew} & threatened\\
            {Bima} & {bhp} & vigorous\\
            {Buginese} & {bug} & wider communication\\
            {Javanese} & {jav} & educational\\
            {Madurese} & {mad} & developing\\
            {Makassarese} & {mak} & threatened\\
            {Minangkabau} & {min} & developing\\
            {Palembang / Musi} & {mui} & wider communication\\
            {Rejang} & {rej} & vigorous\\
            {Sundanese} & {sun} & developing\\
            \bottomrule
        \end{tabular}
    }
    \caption{List of all languages under study in the \datasetname{} benchmark along with their status of language development versus language endangerment.}
    \label{tab:lang-under-study}
\end{table}

In this study, we explore 12 low-resource languages in Indonesian, i.e., Ambon (abs), Batak (btk), Betawi (bew), Bima (bhp), Buginese (bug), Javanese (jav), Madurese (mad), Makassarese (mak), Minangkabau (min), Palembang / Musi (mui), Rejang (rej), and Sundanese (sun). We show the list of all the languages under study in Table~\ref{tab:lang-under-study} and their family tree along with Indonesian (ind) in Figure~\ref{fig:lang-tree}. We provide a more detailed overview of each language in the following paragraphs.

\textbf{Ambonese Malay} (abs) is spoken in various parts of Maluku province. 
It was developed on the island of Ambon in the 16th century, firstly used as (spice) trade language, and now it is used as lingua franca for interethnic communication in the market domain and some media. 
Being a Malay-based creole language, it has around 81\% of lexical similarity with Indonesian. The speakers have marginal intelligibility with Indonesian and difficult intelligibility with North Moluccan Malay~\cite{ethnologue}.
It is written in Latin script. 

\textbf{Batak Toba} (bbc) is spoken in North Sumatra province. 
It is slowly being replaced by Indonesian in urban and migrant areas. It used to be written in the Batak script but is mainly written in Latin script now.

\textbf{Betawi} (bew) is spoken in Tangerang, Banten province, Jakarta, and some cities in West Java province such as Depok, Bekasi, Bogor, and Karawang. 
It is a Malay-based creole distinct from both Indonesian and other Malay-based pidgins and creoles. It was evolved around mid-19th century. 
It functions as a Low variety in a diglossic situation, but has covert prestige when used by the upper class. 
It has unique phonological, morphological, and lexical traits. 
It was influenced by Peranakan Indonesian language and Balinese. 

\textbf{Bima} (bhp) is spoken in Komodo island area in East Nusa Tenggara province and some islands in West Nusa Tenggara province such as Sumbawa island and Banta and Sangeang islands. It has five dialects: Kolo, Sangar, Toloweri, Bima, and Mbojo~\cite{ethnologue}. It is written in Latin script.
 
\textbf{Batak languages} (btk) are a subgroup of the languages of Northwest Sumatra-Barrier Islands spoken by the Batak people in the North Sumatra province and surrounding areas. 
Batak languages can be divided into three groups: Northern, Simalungan, and Southern. The Northern group consists of three languages: Batak Alas-Kluet (btz), Batak Dairi (btd), and Batak Karo (btx). The Simalungan group has one language only, i.e., Batak Simalungun (bts). The Southern group consists of three languages: Batak Angkola (akb), Batak Mandailing (btm), and  Batak Toba (bbc)~\cite{ethnologue}. 
Batak languages are predicate-initial, and have verb systems reminiscent of Philippine languages, although they differ from them in many details~\cite{blust2013austronesian}. 
They were written using the Batak script, but the Latin script is now used for most writing. Our annotators are originating from Batak Toba and Batak Mandailing which are part of the Southern group.

\textbf{Batak Mandailing} (btm) is spoken in North Sumatra (south interior from Padang Sidempuan into Riau) and West Sumatra provinces.
The speakers are shifting to Indonesian in urban and migrant areas~\cite{ethnologue}.
It is written in Batak script.

\textbf{Batak Toba} (bbc) is a language spoken in the North Sumatra province. 
Similarly to Acehnese, it is slowly being replaced by Indonesian in urban and migrant areas. It used to be written in the Batak script but is mainly written in Latin script now.
The Batak languages are verb-initial, and have verb systems reminiscent of Philippine languages, although they differ from them in many details~\cite{blust2013austronesian}.

\textbf{Javanese} (jav) is a language spoken mainly in Java island.
It is the de facto language of provincial identity in central and eastern Java. 
The word order is SVO. It has 21 consonants and 8 vowels. 
It used to be written in Javanese script but since the 20th century, it was mostly written in Latin script.
Javanese differs from most other languages of western Indonesia in contrasting dental and retroflex stops and in the feature of breathy voice or murmur as a phonetic property of its voiced obstruents. Javanese also differs from most languages of the Philippines and western Indonesia in allowing a number of word-initial consonant clusters. It has an elaborate system of speech levels~\cite{blust2013austronesian}.

\textbf{Madurese} (mad) is a language spoken in the East Java province, mainly on Madura Island, south and west of Surabaya city, Bawean, Kangean, and Sapudi islands. 
It has vowel harmony, gemination, rich affixation, three types of reduplication, and SVO basic word order \cite{davies2010grammar}.

\textbf{Makassarese} (mak) is mainly spoken in South Sulawesi province. It has three dialects that form a chain. Those dialects are Lakiung (Gowa), Turatea (Jeneponto), and Bantaeng (Maros-Pangkep). The Gowa dialect is prestigious. 
It has 17 consonants and 5 vowels. The stress is on the penultimate syllable. 
Similar to other Western Malayo-Polynesian languages, it has inclusive and exclusive pronouns, noun head initials, prepositions, definite markers, classifiers, passive markers, and aspect markers~\cite{ethnologue}. 
The speakers, especially young people in the cities, are shifting to Indonesian and Makassar Indonesian. 
It is taught as a subject in primary schools and written in Latin script. The Makassar script is no longer used.

\textbf{Minangkabau} (min) is a language spoken mainly in West Sumatra and other provinces on Sumatra Island such as Bengkulu and Riau. Although it is classified as Malay, it is not intelligible with Indonesian. 
The word order is SVO written in Latin script. 
Standard Minangkabau voice can be characterized as an Indonesian-type system whereas colloquial Minangkabau voice is more effectively characterized as a Sundic-type system~\cite{crouch2009voice}.

\begin{table*}[!t]
\centering
\resizebox{\linewidth}{!}{%
\begin{tabular}{ccccccc}
\toprule
\textbf{Language} & \textbf{\#MT} & \textbf{\#Emotion} & \textbf{\#Sentiment} & \textbf{Split MT} & \textbf{Split Emot} & \textbf{Split Sentiment}\\
\midrule
{btk} & {9,449} & {4,401} & {5,048} & {(6,600, 849, 2,000)} & {(3,000, 401, 1,000)} &  {(3,400, 448, 1,200)}\\
{bew} & {9,449} & {4,401} & {5,048} & {(6,600, 849, 2,000)} & {(3,000, 401, 1,000)} & {(3,400, 448, 1,200)}\\
{sun} & {9,449} & {4,401} & {5,048} & {(6,600, 849, 2,000)} & {(3,000, 401, 1,000)} & {(3,400, 448, 1,200)}\\
{jav} & {9,449} & {4,401} & {5,048} & {(6,600, 849, 2,000)} & {(3,000, 401, 1,000)} & {(3,400, 448, 1,200)}\\
{mad} & {9,449} & {4,401} & {5,048} & {(6,600, 849, 2,000)} & {(3,000, 401, 1,000)} & {(3,400, 448, 1,200)}\\
{mak} & {9,449} & {4,401} & {5,048} & {(6,600, 849, 2,000)} & {(3,000, 401, 1,000)} & {(3,400, 448, 1,200)}\\
{min} & {9,449} & {4,401} & {5,048} & {(6,600, 849, 2,000)} & {(3,000, 401, 1,000)} & {(3,400, 448, 1,200)}\\
{mui} & {1,574} & {733} & {841} & {(1,000, 174, 400)} & {(200, 83, 450)} & {(250, 91, 500)}\\
{rej} & {1,574} & {746} & {828} & {(1,000, 174, 400)} & {(200, 96, 450)} & {(250, 78, 500)}\\
{abs} & {1,574} & {726} & {848} & {(1,000, 174, 400)} & {(200, 76, 450)} & {(250, 98, 500)}\\
{bhp} & {1,579} & {719} & {860} & {(1,000, 179, 400)} & {(200, 69, 450)} & {(260, 100, 500)}\\
\bottomrule
\end{tabular}
}
\caption{Statistics of the NusaTranslation corpus. Split means (training, development, test) respectively.}
\label{tab:nusatranslation_stats}
\end{table*}

\textbf{Musi} (mui) is mainly spoken in South Sumatra province, widespread in the northern two-thirds of the province from the Musi River upstream to Bukit Barisan Mountains and downstream to the coastal swamplands. It is also spoken in the northeast region of Lampung province and around the border areas in Jambi and Bengkulu provinces. 
It has twelve dialects. A mutually intelligible dialect chain stretches along the Musi River with two subgroups: Musi and Palembang. 
The speakers use it for cultural stories and songs, but they prefer Indonesian for educational and religious materials~\cite{ethnologue}.

\textbf{Rejang} (rej) is a language spoken in some parts of Bengkulu and South Sumatra provinces. It has five dialects: Lebong, Kepahiang (Kebanagung), Pasisir, Musi (Curup), and Rawas. Lebong is recognized as a central dialect~\cite{ethnologue}. 
It is written in Latin script and Kaganga script. About 85\% of the speakers live in remote rural areas. Most of them are Muslims.

\textbf{Sundanese} (sun) is a language spoken mainly in the Banten and West Java provinces. It is the de facto language of provincial identity in western Java. The main dialects are Bogor (Krawang), Pringan, and Cirebon. 
It is non-tonal and has 18 consonant and 7 vowel phonemes. The stress is on the penultimate syllable. It has elaborate coding of respect levels. It has been written in Latin script since the middle of the 19th century but was previously written in Arabic, Javanese, and Sundanese scripts.
Sundanese is a predominantly SVO language. It has voice marking and incorporates some (optional) actor-verb agreement, i.e., number and person~\cite{kurniawan2013sundanese}.

\begin{table*}[t]
\centering
\resizebox{0.975\linewidth}{!}{%
\begin{tabular}{ccccccc}
\toprule
\textbf{Language} & \textbf{\#Rhetorical} & \textbf{\#Emotion} & \textbf{\#Topic} & \textbf{Split Rhetorical} & \textbf{Split Emot} & \textbf{Split Topic}\\
\midrule
{btk} & {4,908} & {1941} & {2,125} & {(3,200, 508, 1,200)} & {(1,150, 292, 500)} &  {(1,350, 275, 500)}\\
{bew} & {9,755} & {3,928} & {3,885} & {(6,900, 855, 2,000)} & {(2,700, 430, 800)} & {(2,650, 435, 800)}\\
{bug} & {1,000} & {437} & {443} & {(500, 100, 400)} & {(87, 50,  300)} & {(93, 50, 300)}\\
{jav} & {10,188} & {4,040} & {3898} & {(7,300,  888, 2,000)} & {(2,800, 440, 800)} & {(2,650, 448, 800)}\\
{mad} & {5,211} & {1,762} & {2,867} & {(3,500, 511, 1,200)} & {(1,000, 263, 500)} & {(1,800, 367, 700)}\\
{mak} & {5,471} & {2,303} & {2,576} & {(3,700, 571, 1,200)} & {(1,500, 304, 500)} & {(1,500, 376, 700)}\\
{min} & {8,608} & {3,153} & {3,599} & {(6,000, 608, 2,000)} & {(2,000, 357, 800)} & {(2,400, 399, 800)}\\ 
{mui} & {1,474} & {676} & {648} & {(900, 174, 400)} & {(200, 75, 400)} & {(168, 80, 400)}\\
{rej} & {1,200} & {486} & {505} & {(650, 150, 400)} & {(136, 50,  300)} & {(105, 50, 350)}\\
{sun} & {9,594} & {3,598} & {4,168} & {(6,700, 894, 2,000)} & {(2,400, 400, 800)} & {(2,800, 468, 900)}\\
\bottomrule
\end{tabular}
}
\caption{Statistics of the NusaParagraph corpus. Split means (training, development, test) respectively.}
\label{tab:nusaparagraph_stats}
\end{table*}

\section{Task Description and Statistics of NusaTranslation}
\label{app:nusa_translation}
We developed three tasks from NusaTranslation, i.e., emotion recognition, sentiment analysis, and machine translation. The statistics of the dataset are shown in Table~\ref{tab:nusatranslation_stats}.

\subsection{Emotion Recognition}

From the EmoT~\cite{saputri2018emotion, wilie2020indonlu} part of the translation, develop a parallel emotion recognition dataset of 11 low-resource languages. For most of the languages, we provide new 3,000/401/1,000 train-validation-test splits for all datasets and attach the ids of the original split. In some languages with fewer data, we randomly sample all the original splits and ensure that the test split has a reasonable amount of test samples.

\subsection{Sentiment Analysis}
We develop a parallel sentiment analysis from the IndoLEM Sentiment~ \cite{koto2017inset, koto2020indolem} part of the translation. We provide new 3400/448/1200 train-validation-test splits for most datasets and keep the ids of the original split attached. Similar to the emotion recognition dataset, for the languages with less number of annotators we randomly sample the original splits maintaining a reasonable amount of samples on the test split.

\begin{table}[!t]
    \centering
    \resizebox{0.6\linewidth}{!}{
        \begin{tabular}{cc}
        \toprule
        \textbf{Hyperparams} & \textbf{Values} \\
        \midrule
        {batch size} & {8} \\
        {num epochs} & {100} \\
        {early stop} & {3} \\
        {max norm} & {10} \\
        {optimizer} & {Adam} \\
        {Adam} $\beta$ & {(0.9, 0.999)} \\
        {Adam} $\epsilon$ & {1e-8} \\
        \bottomrule
        \end{tabular}
    }
    \caption{Hyperparameters of pre-trained LMs on machine translation tasks.}
    \label{tab:hyperparameters-translation}
\end{table}

\subsection{Machine Translation} Using the whole constructed translation data, we develop a ind$\leftrightarrow$xxx machine translation task for 11 languages. The scale of our machine translation task is close to one magnitude higher compared to NusaX~\cite{winata2022nusax} which also develops a parallel corpus for 10 local languages spoken in Indonesia.

\section{Task Description and Statistics of NusaParagraph}
\label{app:nusa_writing}

We developed three tasks from NusaParagraph, i.e., rhetoric mode classification, emotion recognition, and topic modeling. The statistics of the dataset is shown in Table~\ref{tab:nusaparagraph_stats}.

\subsection{Rhetoric Mode Classification}
We develop a new rhetoric mode multi-class classification task ranging across 10 low-resource languages. The train-validation-test splits vary between languages as shown in Table~\ref{tab:nusaparagraph_stats}. Paragraphs in the dataset are labeled into one of 5 categories: (1) narrative; (2) argumentative; (3) expository; (4) descriptive; and (5) persuasive.
\subsection{Emotion Recognition}
Using the whole NusaParagraph dataset, we compose an emotion recognition multi-class classification task for 10 low-resource languages. The train-validation-test splits vary between languages as shown in Table~\ref{tab:nusaparagraph_stats}. The emotions expressed in each dataset are labeled into one of 7 emotions: (1) fear; (2) disgusted; (3) sad; (4) happy; (5) shame; (6) angry; and (7) surprise.
\subsection{Topic Modeling}
Each paragraph proposed in the NusaParagraph dataset is annotated with a topic to form a topic modeling task for 10 low-resource languages. The train-validation-test splits vary between languages depending on the dataset size. The details are provided in Table~\ref{tab:nusaparagraph_stats}. Each paragraph is labeled into one of 8 topics: (1) food \& beverages; (2) sports; (3) leisures; (4) religion; (5) culture \& heritage; (6) slice of life; (7) technology; and (8) business.

\begin{table}[!t]
    \centering
    \resizebox{0.6\linewidth}{!}{
        \begin{tabular}{cc}
        \toprule
        \textbf{Hyperparams} & \textbf{Values} \\
        \midrule
        {learning rate} & {1e-5} \\
        {batch size} & {32} \\
        {num epochs} & {100} \\
        {early stop} & {3} \\
        {max norm} & {10} \\
        {optimizer} & {Adam} \\
        {Adam} $\beta$ & {(0.9, 0.999)} \\
        {Adam} $\epsilon$ & {1e-8} \\
        \bottomrule
        \end{tabular}
    }
    \caption{Hyperparameters of pre-trained LMs on classification tasks.}
    \label{tab:hyperparameters-classification}
\end{table}

\begin{table*}[!t]
\centering
\resizebox{\linewidth}{!}{%
\begin{tabular}{cl}
\toprule
{\textbf{No}}&{\textbf{Zero-Shot Prompts}} \\ \midrule
\multicolumn{2}{c}{\cellcolor[HTML]{D9D9D9}\textit{\textbf{Emotion Recognition}}}\\
1 & {[INPUT] => Emotion: [LABELS\_CHOICE]}\\
2 & {Text: [INPUT] => Emotion: [LABELS\_CHOICE]}\\
3 & {[INPUT]\symbol{92}nWhat would be the emotion of the text above? [LABELS\_CHOICE]}\\
4 & {What is the emotion of this text?\symbol{92}nText: [INPUT]\symbol{92}nAnswer: [LABELS\_CHOICE]}\\
5 & {Text: [INPUT]\symbol{92}nPlease classify the emotion of above text. Emotion: [LABELS\_CHOICE]}\\\midrule
\multicolumn{2}{c}{\cellcolor[HTML]{D9D9D9}\textit{\textbf{Rhetoric Mode Classification}}}\\
6 & {[INPUT] => Rhetorical mode: [LABELS\_CHOICE]}\\
7 & {Text: [INPUT] => Rhetorical mode: [LABELS\_CHOICE]}\\
8 & {[INPUT]\symbol{92}nWhat would be the rhetorical mode of the text above? [LABELS\_CHOICE]}\\
9 & {What is the rhetorical mode of this text?\symbol{92}nText: [INPUT]\symbol{92}nAnswer: [LABELS\_CHOICE]}\\
10 & {Text: [INPUT]\symbol{92}nPlease classify the rhetorical mode of above text. Rhetorical mode: [LABELS\_CHOICE]}\\\midrule
\multicolumn{2}{c}{\cellcolor[HTML]{D9D9D9}\textit{\textbf{Topic Modeling}}}\\
11 & {[INPUT] => Topic: [LABELS\_CHOICE]}\\
12 & {Text: [INPUT] => Topic: [LABELS\_CHOICE]}\\
13 & {[INPUT]\symbol{92}nWhat would be the topic of the text above? [LABELS\_CHOICE]}\\
14 & {What is the topic of this text?\symbol{92}nText: [INPUT]\symbol{92}nAnswer: [LABELS\_CHOICE]}\\
15 & {Text: [INPUT]\symbol{92}nPlease classify the topic of above text. Topic: [LABELS\_CHOICE]}\\\midrule
\multicolumn{2}{c}{\cellcolor[HTML]{D9D9D9}\textit{\textbf{Sentiment Analysis}}}\\
16 & {[INPUT] => Sentiment: [LABELS\_CHOICE]}\\
17 & {Text: [INPUT] => Sentiment: [LABELS\_CHOICE]}\\
18 & {[INPUT]\symbol{92}nWhat would be the sentiment of the text above? [LABELS\_CHOICE]}\\
19 & {What is the sentiment of this text?\symbol{92}nText: [INPUT]\symbol{92}nAnswer: [LABELS\_CHOICE]}\\
20 & {Text: [INPUT]\symbol{92}nPlease classify the sentiment of above text. Sentiment: [LABELS\_CHOICE]}\\\midrule
\multicolumn{2}{c}{\cellcolor[HTML]{D9D9D9}\textit{\textbf{Machine Translation}}}\\
21 & {Translate the following text from [SOURCE] to [TARGET].\symbol{92}nText: [INPUT]\symbol{92}nTranslation:}\\
22 & {[INPUT]\symbol{92}nTranslate the text above from [SOURCE] to [TARGET].}\\
23 & {Text in [SOURCE]: [INPUT]\symbol{92}nHow would you translate that in [TARGET]?}\\
24 & {Translate the following [SOURCE] text from to [TARGET].\symbol{92}nText: [INPUT]\symbol{92}nTranslation:}\\
25 & {Text in [SOURCE]: [INPUT]\symbol{92}nText in [TARGET]:}\\\bottomrule
\end{tabular}%
}
\caption{List of prompts used in our zero-shot prompting experiments.}
\label{tab:prompts}
\end{table*}

\section{Experiment Hyperparameters}
\label{app:hyperparameters}

\subsection{Statistical Machine Translation}

For PBSMT, we set the n-gram value of the Moses toolkit to 3. Other parameters were kept to their default values.

\subsection{Neural Machine Translation}
Table~\ref{tab:hyperparameters-translation} shows the hyperparameters used in deep learning models on machine translation experiments in this work. For the learning rate, it follows the configuration of NusaX \cite{winata2022nusax}, while the rest are shown in the following table.

\subsection{Multi-Class Classification}
Table~\ref{tab:hyperparameters-classification} shows the hyperparameters used in deep learning models on classification experiments in this work. Tasks that follow the following parameters include: sentiment analysis, rhetoric mode classification, emotion recognition, and topic modeling. We follow the hyperparameter settings in~\citet{winata2022nusax} that were found to work best.

\section{List of Zero-Shot Prompts}
\label{app:prompts}

We provide the full list of prompts used in our zero-shot prompting experiment in Table~\ref{tab:prompts}.

\end{document}